\documentclass[10pt,twocolumn,letterpaper]{article}

\usepackage[pagenumbers]{cvpr} 

\usepackage{multirow}
\usepackage{xcolor}
\usepackage{colortbl}
\usepackage{utfsym}
\usepackage{tcolorbox}
\usepackage{alltt}

\newcommand{\Method}{Geometrically-Constrained Agent}
\newcommand{\METHOD}{GCA}

\usepackage{amsmath,amsfonts,bm}

\def\eqref#1{equation~\ref{#1}}

\def\1{\bm{1}}

\def\rvx{{\mathbf{x}}}
\def\rvy{{\mathbf{y}}}
\def\rvz{{\mathbf{z}}}

\def\vv{{\bm{v}}}

\DeclareMathAlphabet{\mathsfit}{\encodingdefault}{\sfdefault}{m}{sl}
\SetMathAlphabet{\mathsfit}{bold}{\encodingdefault}{\sfdefault}{bx}{n}

\def\gA{{\mathcal{A}}}

\def\gC{{\mathcal{C}}}

\def\gF{{\mathcal{F}}}

\def\gO{{\mathcal{O}}}

\def\gR{{\mathcal{R}}}

\def\gT{{\mathcal{T}}}

\definecolor{cvprblue}{rgb}{0.21,0.49,0.74}
\usepackage[pagebackref,breaklinks,colorlinks,allcolors=cvprblue]{hyperref}

\def\paperID{xxxx} 
\def\confName{CVPR}
\def\confYear{2026}

\title{Geometrically-Constrained Agent for Spatial Reasoning}

\vspace{-5pt}
\author{
    \vspace{-25pt}\\
    \textbf{
        Zeren Chen$^{1,2*}$,\;
        Xiaoya Lu$^{2,3}$\thanks{Equal contribution. \,\, $^\dag$Corresponding author.}\;\,,\;
        Zhijie Zheng$^{1,2}$,\;
        Pengrui Li$^1$,\;
        Lehan He$^{1,4}$,\;
        Yijin Zhou$^{2,3,4}$
    } \\
    \textbf{
        Jing Shao$^{2}$,\;
        Bohan Zhuang$^{5\dag}$,\;
        Lu Sheng$^{1\dag}$
    } 
    \vspace{4pt} \\
    $^1$School of Software, Beihang University \;\;
    $^2$Shanghai AI Laboratory \\
    $^3$Shanghai Jiao Tong University \;\; 
    $^4$Shanghai Innovation Institute \;\; 
    $^5$ZIP Lab, Zhejiang University \vspace{3pt} \\
    \texttt{\small \{czr1604,lsheng\}@buaa.edu.cn, \{luxiaoya, shaojing\}@pjlab.org.cn} \vspace{6pt}  \\
    Homepage:~\, \url{https://gca-spatial-reasoning.github.io}
    \vspace{-4pt} \\
}

\begin{document}
\maketitle
\begin{abstract}
Vision Language Models (VLMs) exhibit a fundamental semantic-to-geometric gap in spatial reasoning: 
they excel at qualitative semantic inference but their reasoning operates within a lossy semantic space, misaligned with high-fidelity geometry.
Current paradigms fail to bridge this gap.
Training-based methods suffer from an ``oracle paradox,'' learning flawed spatial logic from imperfect oracles.
Tool-integrated methods constrain the final computation but critically leave the VLM's planning process unconstrained, resulting in geometrically flawed plans.
In this work, we propose \textbf{\Method{}} (\textbf{\METHOD{}}), a training-free agentic paradigm that resolves this gap by introducing a formal task constraint. 
Specifically, we strategically decouples the VLM's role into two stages.
First, acting as a semantic analyst, the VLM translates the user's ambiguous query into the formal, verifiable task constraint, which defines the reference frame and objective.
Second, acting as a task solver, the VLM generates and executes tool calls strictly within the deterministic bounds defined by the constraint.
This geometrically-constrained reasoning strategy successfully resolve the semantic-to-geometric gap, yielding a robust and verifiable reasoning pathway for spatial reasoning.
Comprehensive experiments demonstrate that \METHOD{} achieves SOTA performance on multiple spatial reasoning benchmarks, surpassing existing training-based and tool-integrated methods by $\sim$27\%.
\end{abstract}
    
\section{Introduction}
\label{sec:intro}

Intelligent agents operating in real-world applications, such as robotics~\cite{team2025robobrain,ziliotto2025tango,yang2025embodiedbench}, AR/VR~\cite{chandrasegaran2024hourvideo,mangalam2023egoschema,grauman2022ego4d}, and autonomous driving~\cite{tian2025drivevlm,xu2024drivegpt4,fu2024drive}, demand a perceptual understanding of the world akin to humans.
Humans intuitively comprehend their surroundings as a cohesive 3D environment, effortlessly discerning object orientations and complex spatial relationships.
However, equipping Vision Language Models (VLMs) into agents with this holistic \textbf{spatial reasoning} capability remains a critical challenge~\cite{yang2025thinking,yu2025far,yang2025mmsi,jia2025omnispatial,yin2025spatial}.

\begin{figure*}[t]
\begin{center}
\centerline{\includegraphics[width=0.98\linewidth]{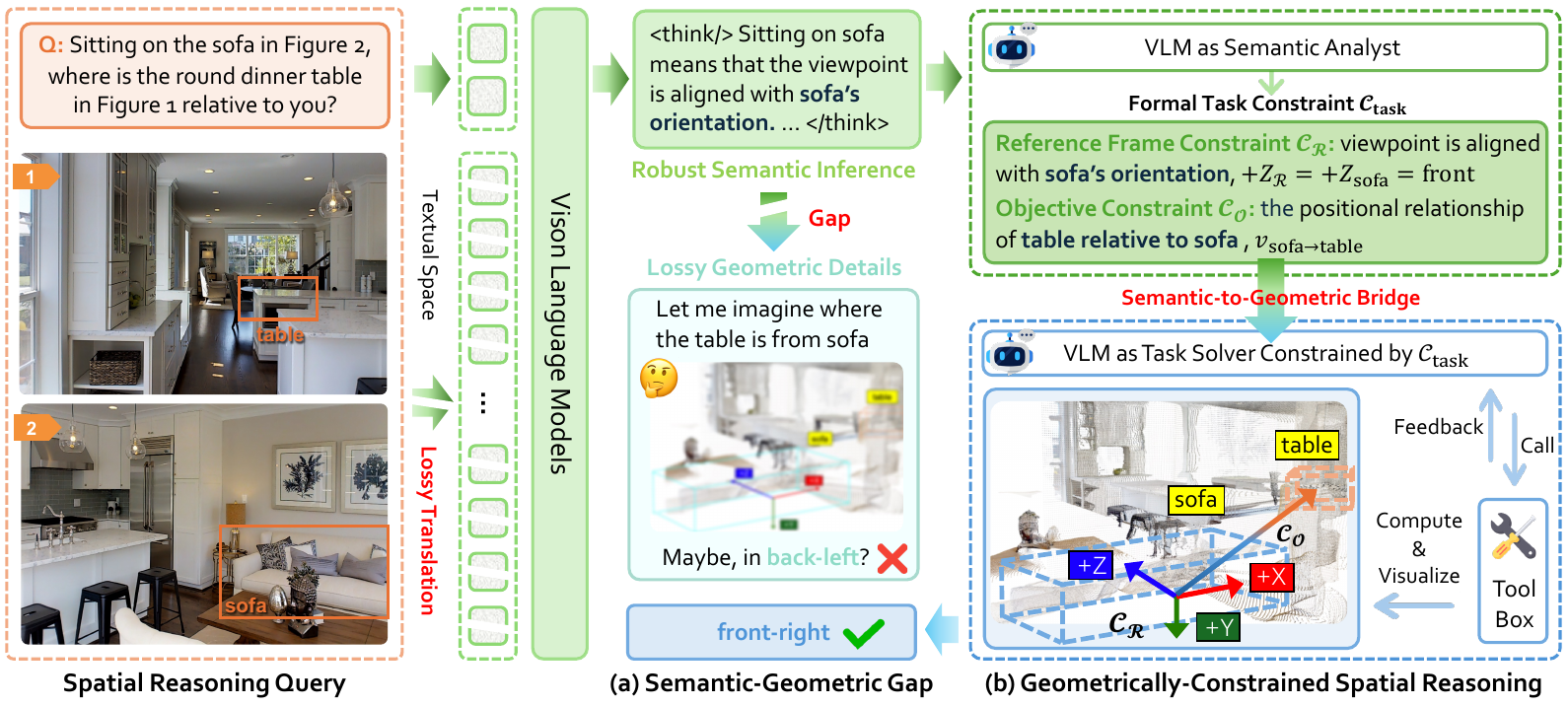}}
\caption{
    \textbf{Overview.}
    \textbf{(a) Semantic-Geometric Gap.} 
    The geometric details required for spatial reasoning are lost when translating visual information into textual space, leading to VLM's flawed reasoning or unconstrained planning. 
    \textbf{(b) Geometrically-Constrained Spatial Reasoning.}
    We propose a formal task constraint that serves as a deterministic bridge between semantics and geometry in spatial reasoning.
}
\label{fig:teaser}
\vspace{-2em}
\end{center}
\end{figure*}

As shown in~\Cref{fig:teaser} (a), current VLMs lossily translate rich visual information into a textual semantic space, leading fine-grained geometric details to be omitted or distorted~\cite{wu2025reinforcing,li2024multimodal}.
This creates a fundamental semantic-to-geometric gap: 
\textit{VLMs excel at probabilistic, qualitative semantic inference, but their lossy semantic space required for spatial reasoning fails to ground high-fidelity geometry.}
For example, a VLM may possess the spatial commonsense (\eg, intuitively knowing that ``sitting on a sofa'' implies a viewpoint aligned with the sofa's orientation), yet critically fail at high-precision geometric computation (\eg, determining the sofa's orientation) and robust spatial imagination (\eg, imagining the user's egocentric perspective).
To reconcile this gap, robust \textbf{constraints} must be imposed, guiding the VLM's reasoning onto a geometrically sound and verifiable pathway.

However, effectively applying these constraints remains a formidable challenge.
Recent approaches that apply implicit constraints via end-to-end training~\cite{ma2025spatialllm,wu2025spatial,li2025spatialladder,ouyang2025spacer,feng2025video,team2025robobrain,wu2025reinforcing,yang2025vlaser} attempt to embed geometric logic by fine-tuning on massive datasets.
These methods, however, face an ``oracle paradox'': 
their data generation relies on oracles like GPT-4o~\cite{hurst2024gpt} which themselves struggle with spatial reasoning~\cite{yang2025thinking,yu2025far,yang2025mmsi,jia2025omnispatial,yin2025spatial}.
Consequently, the VLM is often trained on flawed spatial logic rather than sound geometric principles.
An alternative paradigm, tool integration~\cite{han2025tiger,wu2025spatialscore,ziliotto2025tango}, attempts to bridge this gap by adopting an iterative plan-then-execute strategy, which offloads high-precision geometric computation to deterministic external tools. 
While this constrains the final computation process, the VLM's planning process remains unconstrained.
To plan next step, the VLM must still perform spatial imagination and further decision-making within its lossy semantic space, inevitably producing geometrically flawed plans.
For instance, when asked to reason from the perspective of a user ``sitting on the sofa'' (see~\Cref{fig:teaser}), its unconstrained plan may default to the camera's viewpoint, compromising the problem definition before any tool is even called.

These challenges reveal the critical research question: \textit{How do we bridge the VLM's semantic-to-geometric gap?}
We argue the solution is not to force the VLM to reason about lossy geometric details directly, but to reframe the problem into a task that leverages its inherent strengths: using its spatial commonsense to define a \textbf{formal task constraint} $\gC_\text{task}$ for subsequent computation.
Specifically, this $\gC_\text{task}$ must be 
(1) grammatically rich enough to define complex spatial concepts, such as viewpoints, which elude traditional state-based formalisms
(2) semantically clear enough for a VLM to generate using its qualitative strengths, and 
(3) geometrically sound enough to provide a deterministic, verifiable constraint for subsequent computation.

To this end, we introduce \textbf{\Method{}} (\textbf{\METHOD{}}), a training-free agentic paradigm for geometrically-constrained spatial reasoning.
As shown in~\Cref{fig:teaser} (b), this strategy leverages a formal task constraint, $\gC_\text{task}$, to decouple the reasoning process into two stages:
(1) \textit{Task Formalization.}
The VLM, acting as a semantic analyst, translates the ambiguous query and visual data into the formal, verifiable task constraint $\mathcal{C}_\text{task}$. 
This stage defines what to solve, establishing immutable sub-constraints: a \textbf{reference frame constraint} and an \textbf{objective constraint}.
(2) \textit{Constrained Geometric Computation.} 
The VLM then, acting as a task solver, generates and executes tool calls to compute the final answer, operating strictly within the deterministic bounds defined by $\gC_\text{task}$.
This two-stage decoupling directly bridges the semantic-to-geometric gap.
Through formulating a geometrically sound constraint, we force the VLM to solve deterministic mathematical problems, thereby avoiding the demands for directly computing or imagining about high-fidelity geometric details that are lost in its semantic space.
Extensive experiments demonstrate the effectiveness and generalizability of \METHOD{} paradigm.
%
\METHOD{} yields substantial performance gains when applied to several foundation VLMs (by an average of $\sim$37\%), establishing a new state-of-the-art across a diverse suite of challenging spatial reasoning benchmarks.

\section{Related Work}
\label{sec:related_work}

\begin{figure*}[t]
\begin{center}
\centerline{\includegraphics[width=\linewidth]{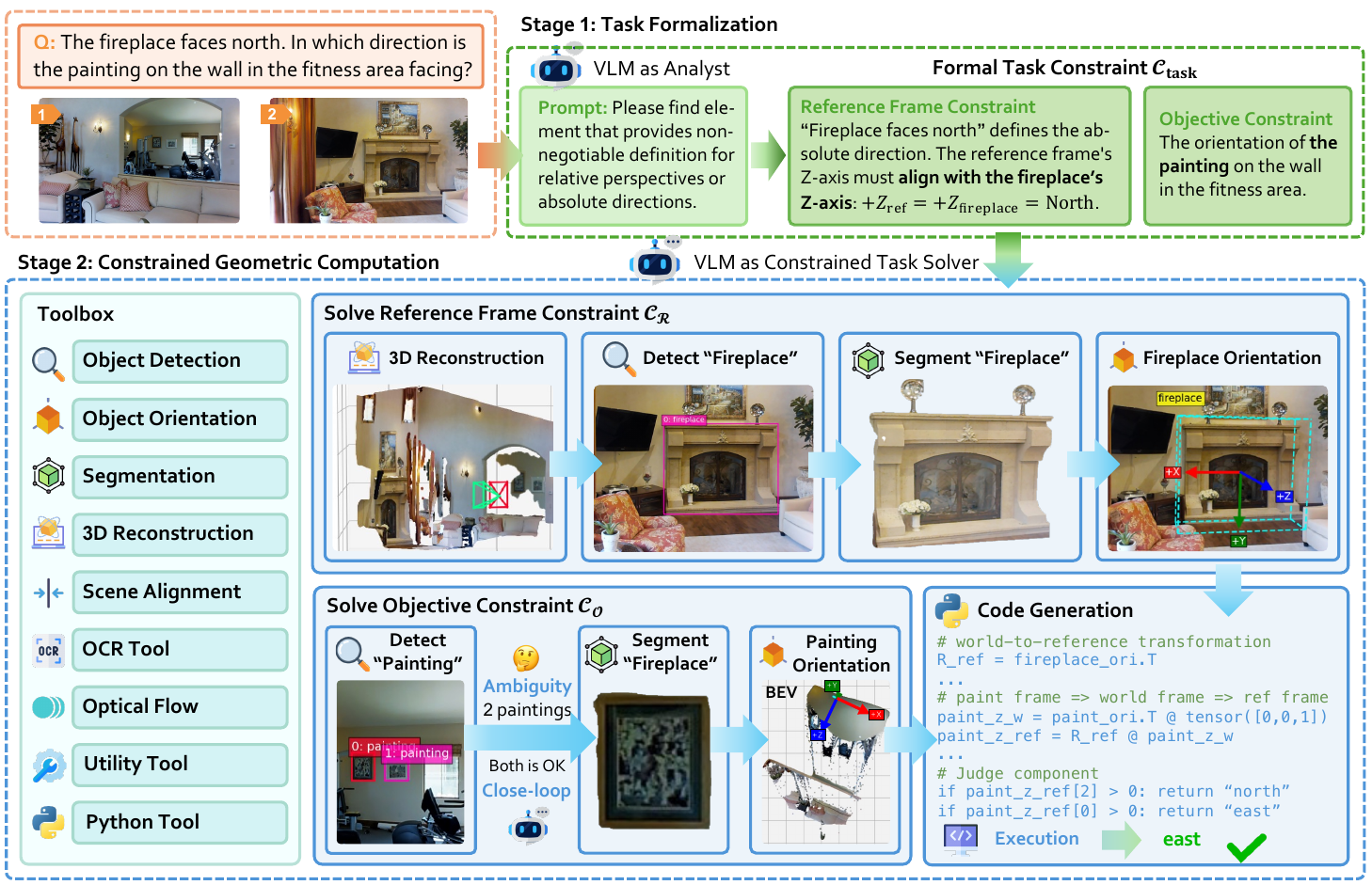}}
\caption{
    \textbf{Overall Paradigm of \METHOD{}.}
    Given a spatial reasoning query, our \METHOD{} leverages a geometrically-constrained reasoning strategy centered on the formal task constraint ($\gC_\text{task}$).
    The VLM first translates the ambiguous query into this explicit $\gC_\text{task}$, establishing a non-negotiable reference frame ($\gC_\gR$) and objective ($\gC_\gO$).
    Strictly constrained by $\gC_\text{task}$, the VLM then orchestrates a toolbox to perform deterministic geometric computation and derive the final answer.
}
\label{fig:overall_paradigm}
\vspace{-2em}
\end{center}
\end{figure*}

\textbf{Spatial Reasoning in Vision Language Models.}
Spatial reasoning, including comprehension and mental manipulation of 3D spatial relationships~\cite{yang2025thinking,yu2025far,yang2025mmsi,yin2025spatial,jia2025omnispatial}, remains a foundational challenge for Vision Language Models (VLMs)~\cite{hurst2024gpt,li2025llava,qwen2025qwen3vl,tong2024cambrian}.
To address this deficit, recent research~\cite{wu2025spatial,feng2025video,yang2025vlaser,li2025spatialladder,ouyang2025spacer,wu2025reinforcing,cheng2024spatialrgpt,ma2025spatialllm,song2025robospatial} focuses on large-scale, end-to-end training on specialized spatial datasets.
These methods attempt to bridge the 2D-3D cognitive gap by incorporating geometric priors, such as explicit 3D structural features~\cite{wu2025spatial}, or depth maps~\cite{cheng2024spatialrgpt}, directly into the VLM's architecture, but they are often hindered by the reliance on high-quality datasets generated by flawed oracle.
Another line of research introduce tool-integrated reasoning~\cite{han2025tiger,ziliotto2025tango,ma2024spatialpin,wu2025spatialscore,lin2025embodiedcoder} to offloads deterministic geometric computation to external modules.
For example, SpatialAgent~\cite{wu2025spatialscore} and TIGeR~\cite{han2025tiger} focus on translating the input query directly into an iterative sequence of tool executions.
However, unconstrained planning process could lead to geometrically-flawed results, causing the agent to conflate ``what to solve'' with ``how to solve it''.

\vspace{+0.5mm}
\noindent\textbf{Constrained-Guided Reasoning.}
Constraint-guided reasoning, which involves restricting a search space by defining variables and the constraints governing them~\cite{russell1995modern}, has been adapted to manage the probabilistic nature of LLMs and VLMs.
A primary application is in neuro-symbolic reasoning~\cite{hitzler2022neuro,pan2023logic,ye2023satlm,hao2024planning,bhuyan2024neuro,sultan2025towards}, where the LLM is constrained to act as a translator, converting ambiguous natural language into a formal, verifiable representation.
For example, LogicLM~\cite{pan2023logic} leverage LLMs to translate NL problems into task-specific formalisms like formal logic.
This constraint-guided reasoning can also be extended to planning~\cite{liu2023llmp,yang2025guiding,chen2024rh20t,huang2025rekep,pan2025omnimanip}.
Frameworks like LLM+P~\cite{liu2023llmp} uses an LLM to translate an NL problem into a formal PDDL format and then applies an optimal planner to generate the plan.
Similarly, ReKep~\cite{huang2025rekep} employs a VLM to translate a free-form language into relational keypoint constraints and solves the constraints to generate final robot actions.

\section{Methodology}
\label{sec:method}

As illustrated in~\Cref{fig:overall_paradigm}, we propose \Method{} (\METHOD{}), a training-free agentic paradigm designed for geometrically-constrained spatial reasoning.
The core of \METHOD{} is the introduction of a formal task constraint $\gC_\text{task}$ that serves as a deterministic bridge between semantics and geometry.
\Cref{subsec:cgsr} defines this geometrically-constrained paradigm.
\Cref{subsec:formalize} details the formal task constraint $\gC_\text{task}$ and its automated generation.
\Cref{subsec:compute} describes the subsequent constrained computation stage, which is strictly governed by this constraint.
Finally, \Cref{subsec:discussion} discusses how \METHOD{} resolves the VLM's semantic-to-geometric gap in spatial reasoning.

\subsection{Geometrically-Constrained Spatial Reasoning}
\label{subsec:cgsr}

Contemporary agentic frameworks often model reasoning as a generic, iterative policy.
Those based on the ReAct framework~\cite{yao2022react}, for example, can be defined by:
\begin{equation}
    r_t=\gA(q,\vv,\gT,r_{t-1}).
\end{equation}
In this framework, an agent $\gA$ produces a response $r_t$ based on a query $q$, visual information $\vv$, a set of tools $\gT$, and its past history $r_{t-1}$. 
This generic policy $\gA$ is unconstrained, making it unreliable for high-stakes, deterministic domains like spatial reasoning.
Recent work~\cite{han2025tiger,wu2025spatialscore} attempts to mitigate this by using external tools to constrain the final computation.
However, they fail to constrain the VLM's planning process.
The VLM may still rely on its flawed spatial imagination in lossy semantic space to formulate plan, conflating ``what to solve'' with ``how to solve it''.

We solve this by replacing the generic policy $\gA$ with a two-stage process.
This paradigm is built on the formal task constraint $\gC_\text{task}$, which functions as the architectural scaffolding to align the VLM's asymmetric capabilities:
\begin{equation}
\begin{gathered}
    \gC_\text{task} \leftarrow \gF_\text{formalize}(q,\vv), \\
    r_t = \gF_\text{compute}(\gC_\text{task}, \gT, r_{t-1}).
\end{gathered}
\end{equation}
In the $\gF_\text{formalize}$ stage, the VLM acts as a semantic analyst, translating the ambiguous query $q$ and visual information $\vv$ into a formal, verifiable task constraint $\gC_\text{task}$.
This stage defines what to solve by establishing the necessary geometric scaffolding (\eg{}, the reference frame and target subjects).
In the $\gF_\text{compute}$ stage, the VLM's role shifts to a task solver.
Governed by the constraint $\gC_\text{task}$ established in the $\gF_\text{formalize}$ stage, the VLM iteratively executes tool calls to acquire necessary geometric data and perform final computations.

\begin{figure}[t]
\begin{center}
\centerline{\includegraphics[width=\linewidth]{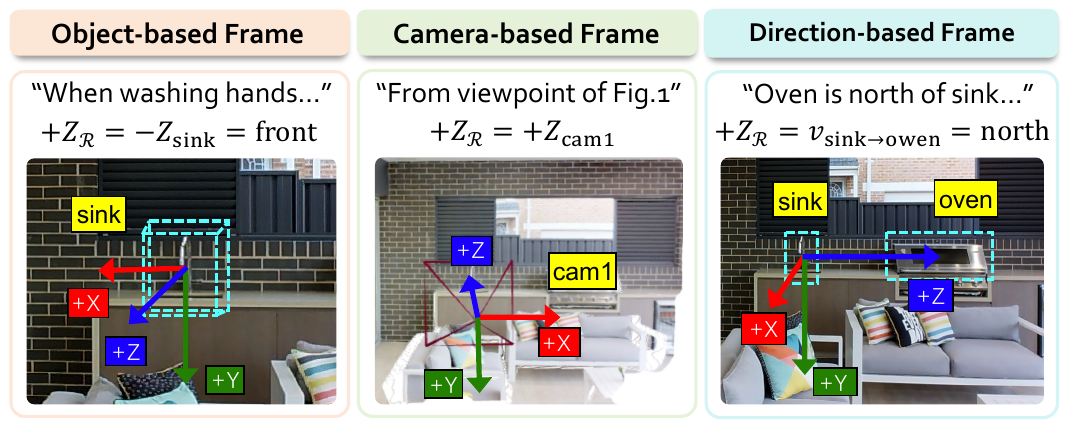}}
\caption{
    \textbf{Reference Frame.}
    Here, $v_{\text{sink}\rightarrow\text{owen}}$ denotes a vector calculated by ``$\text{normalize}\left(\text{Centroid(owen)}-\text{Centroid(sink)}\right)$''.
}
\label{fig:ref_frame}
\vspace{-2.5em}
\end{center}
\end{figure}

\subsection{Task Constraint Formalization}
\label{subsec:formalize}

\subsubsection{Constraint for Spatial Reasoning}
\label{subsubsec:task_const}

While existing constraint-guided reasoning paradigms leverage formalisms such as PDDL~\cite{liu2023llmp} or relational keypoint constraints~\cite{huang2025rekep}, these constraints are insufficient for spatial reasoning. 
PDDL, for instance, excels at describing discrete, symbolic object states (\eg{}, ``\texttt{is\_on(A, B)}'') but fundamentally lacks the geometric grammar to express the continuous, relative, and perspective-dependent nature of spatial queries (\eg{}, egocentric \emph{vs.} allocentric viewpoints).
This gap necessitates a new formalism.
We thus propose a novel \textbf{formal task constraint} $\gC_\text{task}$ specifically designed to capture the geometric nature of spatial reasoning.
We define $\gC_\text{task}$ as a tuple containing two key sub-constraints: 
a single, non-negotiable \textbf{Reference Frame Constraint} ($\gC_\gR$) that defines the coordinate system for answering the query, 
and an \textbf{Objective Constraint} ($\gC_\gO$) that specify the objective to be measured within that frame.

\vspace{+0.5mm}
\noindent\textbf{Reference Frame Constraint.}
Humans intuitively understand spatial language (\eg{}, ``north of'') by grounding it within a specific coordinate system, namely a reference frame ($\gR$).
In contrast, VLM failures often stem from ambiguity in this crucial grounding step, causing them to adopt flawed geometrically flawed plans~\cite{yang2025mmsi} (\eg{}, defaulting to the camera's viewpoint).
The $\gF_\text{formalize}$ stage addresses this ambiguity by requiring the VLM to first formally anchor $\gR$ to the scene's geometry.

We model all spatial queries as requiring a 3D cartesian coordinate system defined by an origin $O_\gR$ and three orthogonal basic vectors ($\rvx_\gR, \rvy_\gR, \rvz_\gR$).
This system follows the OpenCV convention, where $+\rvz_\gR$ points forward, $+\rvy_\gR$ points down and $+\rvx_\gR$ follows the right-hand rule.
The agent's task is to anchor $\gR$ to one of three geometric primitives (see~\Cref{fig:ref_frame}) derived from the visual information:
\vspace{+0.5mm}
\begin{itemize}
    \item \textbf{Object-based Frame.}
    $\gR$ is defined by an object's intrinsic coordinate system. 
    For example, the query ``when the user is washing hand'' implies a reference frame defined by $+\rvz_\gR=-\rvz_\text{sink}$ (one must face a sink to wash hand). 
    \vspace{+0.5mm}
    \item \textbf{Camera-based Frame.}
    $\gR$ is defined by a specific camera's viewpoint.
    For ``from the viewpoint of Figure 1'', the reference frame is defined by $+\rvz_\gR=+\rvz_\text{cam1}$.
    \vspace{+0.5mm}
    \item \textbf{Direction-based Frame.}
    $\gR$ is defined by a vector connecting two locations.
    For ``Owen is north of sink'', the reference frame is defined by $+\rvz_\gR=\text{normalize}\left(\text{Centroid(owen)}-\text{Centroid(sink)}\right)=\text{north}$.
\end{itemize}
\vspace{+0.5mm}
The output of this step is a human-readable and machine-parsable definition of $\gR$, which becomes a non-negotiable constraint $\gC_\gR$ for all subsequent computation.

\vspace{+0.5mm}
\noindent\textbf{Objective Constraint.}
Concurrently, the agent identifies the objective $\gO$ from the query. 
This constraint $\gC_\gO$ defines \textit{what} must be measured relative to the established $\gR$.
For the query, ``Is chair to the west of toaster?'', the toaster defines $\gC_\gR$, while the positional relationship between toaster and chair is the objective constraint $\gC_\gO$.

\subsubsection{Automated Formalization via VLM}
\label{subsubsec:formalize_impl}

We exploit the VLM's innate strength in semantic interpretation to generate $\gC_\text{task}$ automatically.
Acting as a semantic analyst, the VLM performs qualitative interpretation, guided by the formal definitions of $\gC_\text{task}$, to generate the $\gC_\text{task} = (\gC_\gR, \gC_\gO)$.
This formal task constraint, generated by the VLM but grounded in geometry, serves as the geometrically sound contract for the $\gF_\text{compute}$ stage.
In our implementation, we enforce this architectural decoupling procedurally. 
The VLM executes the $\gF_\text{formalize}$ stage and formalizes the $\gC_\text{task}$ before any computation begins.

\subsection{Constrained Geometric Computation}
\label{subsec:compute}

\subsubsection{Tool Integration and Code Generation}
\label{subsubsec:tool_integrate}

Once the formal task constraint $\gC_\text{task} = (\gC_\gR, \gC_\gO)$ is established, the VLM's role shifts to a constrained task solver.
This $\gF_\text{compute}$ stage then operates as a ReAct-style framework, consuming the $\gC_\text{task}$ as an immutable constraint.
This execution is not a one-shot generation but an iterative, closed-loop process involving data acquisition, ambiguity resolution, and augmented computation.

\vspace{+0.5mm}
\noindent\textbf{Data Acquisition.}
$\gC_\text{task}$ dictates a set of geometric ingredients that the agent must acquire.
For instance, as shown in~\Cref{fig:overall_paradigm}, to instantiate an object-based frame $\gR$ defined by a sink, the agent must acquire the orientation of that sink.
The $\gF_\text{compute}$ stage begins by generating a sequence of tool calls to parameterize the geometry, and acquire all variables necessary to instantiate $\gC_\text{task}$.

\vspace{+0.5mm}
\noindent\textbf{Tool Orchestration and Ambiguity Resolution.}
The VLM is responsible for managing tool feedback and resolving ambiguity, ensuring the data acquired from tools correctly binds to the symbols in $\gC_\text{task}$.
For example, considering $\gC_\gO$ involves an object like ``leftmost chair'', the perception tool returns several ``chair'' detections.
The VLM analyzes this feedback (\eg{}, visualizing bounding boxes) and resolves the ambiguity by determining which object index correctly corresponds to the context (``leftmost'') specified. 
This closed-loop mechanism allows the agent to handle noisy tool outputs while ensuring the final computation remains strictly grounded in the intent of $\gC_\text{task}$.

\vspace{+0.5mm}
\noindent\textbf{Knowledge-Augmented Code Generation.}
Once all variables in $\gC_\text{task}$ are bound to concrete geometric data, the agent invokes a code generator for the final computation.
To prevent the coder from hallucinating incorrect formulas, we leverage a knowledge-augmented strategy, which functions analogously to a static Retrieval-Augmented Generation (RAG)~\cite{lewis2020retrieval,gao2023retrieval} system.
Specifically, when invoking the code generator, the VLM specifies a high-level requirement and the necessary bound variables (\eg{}, object's orientation).
Instead of expecting the coder to generate complex geometric formulas from memory, our framework maintains a pre-prepared, fixed library of basic, verified geometric formulas. 
Based on the data types of the bound variables, the system automatically retrieves the relevant, fixed set of formulas (\eg{}, object's local-to-world transformation formula) and injects them directly into the code generator's context.
This ensures the computation steps do not produce black-box guesses, but rather deterministic results, derived from a formally structured task and sound geometric principles.
%
More details are provided in~\Cref{sec:app_prompt}.

\subsubsection{Toolbox}
\label{subsubsec:toolbox}

We equips the agent with perceptual and computation capabilities required to execute its geometrically-constrained reasoning flow in $\gF_\text{compute}$, as shown in~\Cref{fig:overall_paradigm}.
%
Detailed APIs for all tools are provided in~\Cref{subsec:app_tool_interfaces}.

\vspace{+0.5mm}
\noindent\textbf{Geometry and Perception Tools.}
These tools are responsible for parameterizing the visual world.
%
``3D Reconstruction'' tool leverages foundational models like VGGT~\cite{wang2025vggt} to build a unified, high-fidelity 3D representation of the scene.
%
This provides the geometric context required for complex scenarios.
This category also contains a suite of 2D perception tools, such as ``Object Detection'' for open-vocabulary object detection, ``Segmentation'' for instance segmentation.

\begin{table*}[t]
\caption{
    \textbf{
        Experimental Results on Several Spatial Reasoning Benchmarks.
    }
    The best and second best results are shown in \textbf{bold} and \underline{underlined}, respectively.
    ``\textbf{Avg.}'' denotes the average of overall accuracy across all benchmarks.
    More details about these benchmarks' subcategory (\eg{}, ``PR.'') are provided in Appendix.
}
\vspace{-1.5em}
\label{tab:main_results}
\begin{center}
\resizebox{\linewidth}{!}{%
\begin{tabular}{@{}lccccccccccccccccccc}
    \toprule
    &
    \multicolumn{5}{c}{\textbf{MMSI-Bench}} &
    \multicolumn{4}{c}{\textbf{MindCube-tiny}} &
    \multicolumn{3}{c}{\textbf{OmniSpatial}} &
    \multicolumn{3}{c}{\textbf{SPBench}} &
    \multicolumn{3}{c}{\textbf{CV-Bench}} &
    \multirow{2}{*}{\vspace{+0mm}\textbf{Avg.}} \\
    \cmidrule(lr){2-6}\cmidrule(lr){7-10}\cmidrule(lr){11-13}\cmidrule(lr){14-16}\cmidrule(lr){17-19}
    & 
    PR. & Attr. & Mot. & MSR & All &
    Rot. & Ard. & Amg. & All & 
    Dyn. & Pers. & All &
    SI & MV & All &
    2D & 3D & All \\
    \midrule
    \rowcolor{gray!10}\multicolumn{20}{l}{\textbf{\textit{Baseline Foundation VLMs}}} \vspace{+0.7mm} \\
    \hspace{+1mm} Qwen3-VL-Thinking~\cite{qwen2025qwen3vl} & 33.7 & \underline{40.0} & 23.3 & 31.8 & 32.6 & \underline{87.0} & 47.3 & 35.0 & 47.3 & 60.5 & 43.9 & 51.0 & 51.9 & 61.2 & 54.1 & \underline{81.9} & \underline{92.6} & \underline{86.8} & 54.4 \\
    \hspace{+1mm} GLM-4.5V~\cite{hong2025glm}              & 35.6 & 36.9 & 29.3 & 30.3 & 33.8 & 60.0 & 25.5 & 42.2 & 39.6 & 58.6 & \underline{47.2} & 52.1 & 50.0 & 55.1 & 51.3 & 80.7 & 91.6 & 85.6 & 52.5 \\
    \hspace{+1mm} GPT-4o~\cite{hurst2024gpt}               & 28.0 & 32.3 & \underline{36.0} & 30.8 & 30.3 & 33.5 & 35.0 & 37.2 & 35.8 & 58.7 & 46.2 & 51.5 & 42.4 & 48.3 & 43.8 & 69.4 & 84.9 & 76.5 & 47.6 \\
    \hspace{+1mm} Gemini-2.5-Pro~\cite{comanici2025gemini} & \underline{39.0} & 36.2 & 33.3 & \underline{34.3} & \underline{36.9} & \textbf{89.5} & \underline{54.5} & \underline{48.8} & \underline{57.5} & \underline{70.7} & 44.6 & \underline{55.8} & 55.6 & 58.3 & 56.3 & 81.2 & 92.5 & 86.3 & \underline{58.5} \\
    \midrule
    \rowcolor{gray!10}\multicolumn{20}{l}{\textbf{\textit{Training-based Spatial VLMs}}} \vspace{+0.7mm} \\
    \hspace{+1mm} SpatialLLM~\cite{ma2025spatialllm}       & 24.5 & 23.1 & 22.7 & 30.8 & 25.3 & 34.0 & 26.8 & 33.0 & 31.1 & 59.6 & 42.9 & 49.5 &32.2 &26.4 &30.7 & 51.3 & 78.6 & 64.5 & 40.2 \\
    \hspace{+1mm} Spatial-MLLM~\cite{wu2025spatial}        & 28.5 & 25.4 & 18.0 & 26.3 & 26.1 & 33.8 & 34.5 & 28.3 & 32.1 & 37.2 & 42.1 & 40.0 &52.0 & 52.0 & 52.0 & 59.5 & 63.3 & 61.2 & 42.3 \\
    \hspace{+1mm} SpatialLadder~\cite{li2025spatialladder} & 30.3 & 23.3 & 16.0 & 21.2 & 25.4 & 30.5 & 39.8 & 47.8 & 42.3 & 46.5 & 43.1 & 44.5 &\textbf{70.2} & \textbf{70.9} & \textbf{70.3} & 72.4 & 74.9 & 73.7 & 51.2 \\
    \hspace{+1mm} SpaceR~\cite{ouyang2025spacer}          & 29.1 & 29.4 & 21.9 & 22.5 & 26.9 & 29.8 & 30.0 & 26.8 & 28.3 & 53.5 & 40.5 & 46.0 & 48.6 & 59.4 & 51.1 & 74.1 & 77.4 & 75.6 & 45.7 \\
    \hspace{+1mm} Video-R1~\cite{feng2025video}            & 30.5 & 25.4 & 22.0 & 26.8 & 27.8 & 30.0 & 30.5 & 41.3 & 35.8 & 50.0 & 44.2 & 46.7 & 44.8 & 40.7 & 43.8 & 73.5 & 74.7 & 74.0 & 45.6 \\
    \hspace{+1mm} RoboBrain-2.0~\cite{team2025robobrain}   & 28.9 & 28.8 & 22.5 & 28.0 & 28.9 & 29.7 & 35.8 & 45.2 & 39.6 & 49.4 & 42.2 & 45.2 & 49.1 &46.8 & 48.5 & 77.1 & 90.7 & 83.4 & 49.1 \\
    \hspace{+1mm} VILASR~\cite{wu2025reinforcing}          & 35.9 & 26.0 & 21.0 & 23.2 & 29.8 & 34.4 & 25.7 & 29.4 & 29.1 & 37.5 & 42.2 & 40.2 & 50.2 & 57.6 & 51.9 & 75.7 & 77.7 & 76.6 & 45.5 \\
    \hspace{+1mm} VLaser~\cite{yang2025vlaser}             & 29.8 & 26.9 & 26.0 & 18.9 & 27.3 & 31.5 & 24.8 & 38.2 & 32.6 & 39.1 & 42.6 & 41.1 &53.2 &\underline{69.2} &56.9 & 79.9 & 87.8 & 83.6 & 48.3 \\
    \midrule
    \rowcolor{gray!10}\multicolumn{20}{l}{\textbf{\textit{Tool-Integrated Spatial Agents}}} \vspace{+0.7mm} \\
    \hspace{+1mm} TIGeR~\cite{han2025tiger}                & 29.1 & 27.7 & 26.0 & 25.8 & 27.8 & 33.0 & 28.3 & 26.7 & 28.3 & 52.9 & 45.7 & 49.8 & 48.7 & 38.8 & 46.3 & 75.2 & \textbf{95.7} & 84.5 & 47.3 \\
    \midrule
    \hspace{+1mm} \METHOD{} (ours)                         & \textbf{52.8} &  \textbf{45.0} & \textbf{44.7} & \textbf{38.0} & \textbf{47.6} & 82.0 & \textbf{61.8} & \textbf{59.8} & \textbf{64.2} & \textbf{73.6} & \textbf{58.6} & \textbf{65.1} & \underline{61.7} & 61.9 & \underline{61.8} & \textbf{83.6} & 90.8 & \textbf{86.9} & \textbf{65.1} \\
    \bottomrule
\end{tabular}
}
\end{center}
\vspace{-1em}
\end{table*}

\vspace{+0.5mm}
\noindent\textbf{Computation and Utility Tools.}
These tools operate on the data extracted by the perception tools and executes the final deterministic geometric computation.
``Python Tool'' is the core computational engine, which prompts the VLM to generate and execute Python code in a sandbox environment, using the knowledge-augmented strategy.
This category also includes essential utilities (``Utility Tool'').
For example, ``\texttt{project\_box\_to\_points}'' bridges 2D perception to 3D computation by converting 2D bounding boxes into corresponding 3D point clouds.

\subsection{Discussion}
\label{subsec:discussion}

Our \METHOD{} decouples VLM's spatial reasoning through the formal constraint $\gC_\text{task}$, jointly addressing two core deficiencies in spatial reasoning.

\vspace{+0.5mm}
\noindent\textbf{$\gF_\text{formalize}$ Solves Flawed Planning and Imagination.}
Directly solving an ambiguous query forces the VLM to plan and perform spatial imagination within its native lossy semantic space.
This is a primary failure mode, as unconstrained planning can lead to geometrically flawed assumptions before any computation even begins.
Our paradigm resolves this by reframing the problem.
Leveraging VLM's strength in qualitative semantic interpretation, the $\gF_\text{formalize}$ stage transform the original spatial query into a deterministic mathematical problem with constraint, preventing the VLM to solve the query in its lossy semantic space directly.

\vspace{+0.5mm}
\noindent\textbf{$\gF_\text{compute}$ Solves Flawed Execution and Computation.}
In this stage, the VLM acting as the task solver, orchestrating external tools to execute the plan.
Crucially, its entire reasoning and execution process is bound by the formal task constraint $\gC_\text{task}$ generated in $\gF_\text{formalize}$.
This ensures that all subsequent high-precision computations are executed strictly within the deterministic, geometrically sound constraint, effectively bridging the semantic-to-geometric gap.

\section{Experiments}
\label{sec:exp}

\subsection{Experimental Setup}
\label{subsec:exp_setup}

\textbf{Implementation Details.}
\METHOD{} is implemented as a training-free agentic paradigm, requiring no model fine-tuning.
It centers on a VLM responsible for both stages of our paradigm: 
acting as the semantic analyst to generate the $\gC_\text{task}$ in the $\gF_\text{formalize}$ stage, 
and as the task solver to manage a suite of off-the-shelf foundation models for perception and computation~\cite{wang2025vggt,wang2025moge,liu2024grounding,ravi2024sam,wang2024orient}.
For our primary experiments, we utilize Qwen3-VL-Thinking~\cite{qwen2025qwen3vl} as the central VLM.
To assess the paradigm's generalizability, we also evaluate other leading VLMs in our ablation studies, including GLM-4.5V~\cite{hong2025glm}, GPT-4o~\cite{hurst2024gpt}, and \etc{}
All open-source VLMs are deployed using the vLLM inference engine~\cite{kwon2023efficient} for efficiency.
The agent's architecture is built using Ray~\cite{moritz2018ray} for concurrent tool execution and LangGraph for robust state management.

\vspace{+0.5mm}
\noindent\textbf{Evaluation Benchmarks and Counterparts.}
We conduct comprehensive experiments on several spatial reasoning benchmarks. 
As our current toolbox is primarily designed for image-based inputs, we focus on evaluations that test complex spatial logic from single and multiple images, including MMSI-Bench~\cite{yang2025mmsi}, MindCube-tiny~\cite{yin2025spatial}, OmniSpatial (Perspective Taking + Dynamic Reasoning)~\cite{jia2025omnispatial}, SPBench~\cite{li2025spatialladder} and CV-Bench~\cite{tong2024cambrian}.
For all benchmarks, we report both overall accuracy (\%) and subcategory accuracy (\%).
We compare our paradigm against several counterparts, including baseline foundation VLMs~\cite{hurst2024gpt,comanici2025gemini,qwen2025qwen3vl,hong2025glm}, training-based methods~\cite{ma2025spatialllm,wu2025spatial,li2025spatialladder,ouyang2025spacer,feng2025video,team2025robobrain,wu2025reinforcing,yang2025vlaser} and tool-integrated agents~\cite{han2025tiger}.

\subsection{Main Results}
\label{subsec:main_results}

\noindent\textbf{SOTA Performance.}
As shown in~\Cref{tab:main_results}, \METHOD{} establishes a new state-of-the-art across a wide range of spatial reasoning benchmarks, achieving an average accuracy of 64.8\%.
Our geometrically-constrained paradigm surpasses the strongest foundation VLM baseline (Gemini-2.5-Pro~\cite{comanici2025gemini} by 12\%) and demonstrates a massive lead over other training-based (\eg{}, SpatialLadder~\cite{li2025spatialladder} by 27\%) or agentic approaches (\eg{}, TIGeR~\cite{han2025tiger} by 38\%).
These results strongly validate that our strategy, centered on the $\gC_\text{task}$, successfully bridges the VLM's semantic-to-geometric gap.

\vspace{+0.5mm}
\noindent\textbf{Effectiveness on Challenging Benchmarks.} 
The advantage of our constrained paradigm is most pronounced on complex, multi-step spatial reasoning benchmarks. 
For example, on MMSI-Bench, the performance of even SOTA foundation VLMs remain severely limited.
Considering its 4-choice questions, most counterparts perform near the 25\% random-guess threshold.
In contrast, \METHOD{} achieves an overall accuracy of 47.6\%, surpassing the strongest VLM baseline (Gemini-2.5-Pro) by a 28\% relative improvement.
A similar trend is evident on other challenging benchmarks like MindCube-tiny, where \METHOD{} (64.2\%) also significantly outperforms the top baselines.
This superior performance stems directly from our paradigm. 
The introduction of $\gC_\text{task}$ prevents the VLM from defaulting to flawed semantic shortcuts or falling into a lossy spatial imagination.

\vspace{+0.5mm}
\noindent\textbf{Generalizability Across Benchmarks.}
Our training-free paradigm also demonstrates superior generalizability compared to training-based specialists, which often suffer from biases inherent to their training data.
For example, SpatialLadder~\cite{li2025spatialladder} is fine-tuned on data originating from the same source as the SPBench, leading to a high in-domain score of 70.3\%.
However, its performance on out-of-domain benchmarks is suboptimal, where \METHOD{} consistently outperforms it, often by a margin of $\sim$20 points.
A similar bias affects TIGeR~\cite{han2025tiger}.
While its tools theoretically support multi-view processing, the model is primarily trained on single-image tasks.
Consequently, it performs well on single-image benchmarks like CV-Bench but fails on multi-view benchmarks such as MMSI-Bench and MindCube.
\METHOD{}, in contrast, is not compromised by these training priors and leverages its multi-view tools as dictated by the problem.
This demonstrates that our \METHOD{}, which forces the VLM to derive a geometrically sound task constraint for each new problem, provides a more robust and generalizable pathway to spatial reasoning.

\subsection{Ablation Study}
\label{subsec:ablation}

In this section, we conduct extensive ablation studies to dissect the \METHOD{} paradigm and validate its core design. 
Our analysis aims to answer four critical questions.
\textbf{(1)} How essential is the formal task constraint $\gC_\text{task}$?
\textbf{(2)} How generalizable is the \METHOD{} paradigm across different VLMs?
\textbf{(3)} What is the contribution of each system component?

\begin{figure}[t]
\begin{center}
\centerline{\includegraphics[width=0.87\linewidth]{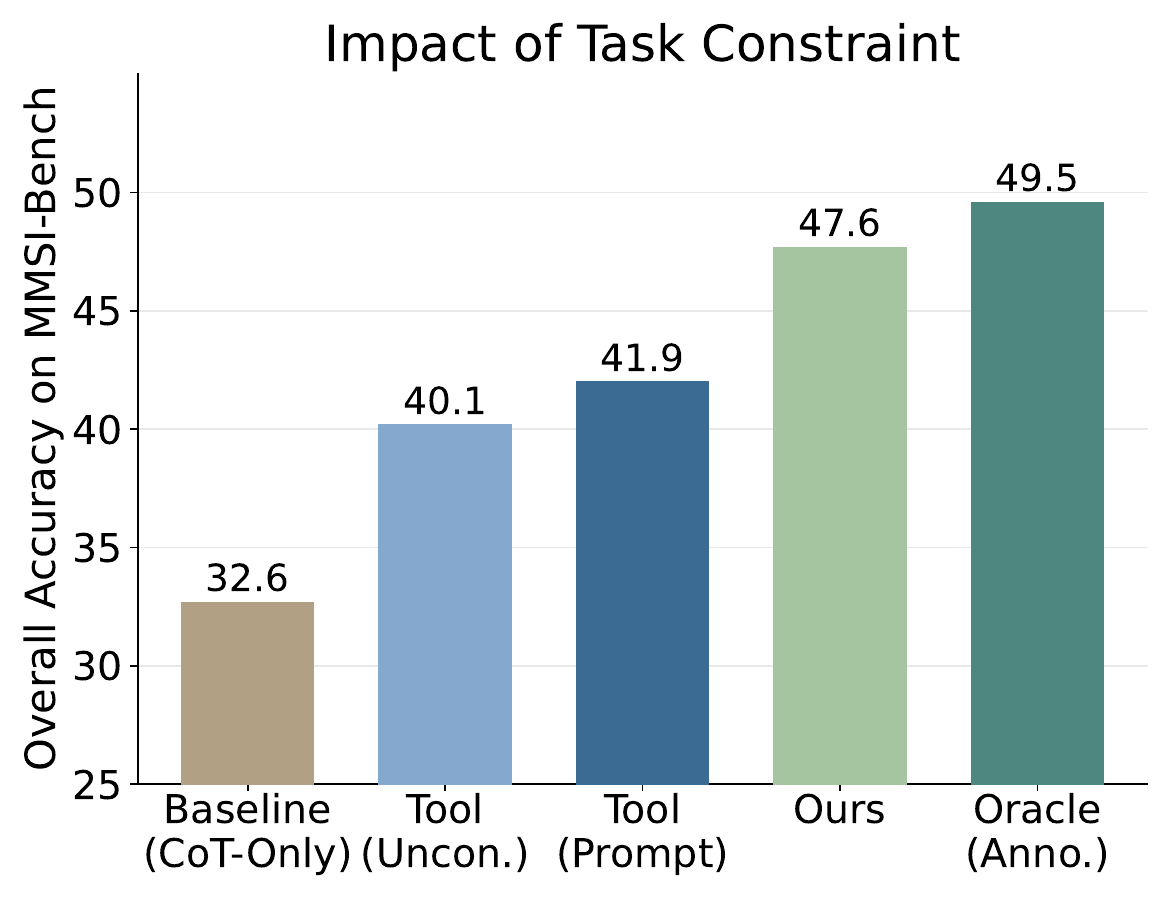}}
\caption{
    \textbf{Ablation Study on Formalization.}
    We compare our method in against several baselines: 
    (1) no tool integration (``Baseline (CoT-Only)''),
    (2) unconstrained tool integration with (``Tool (Prompt)'') or without (``Tool (Uncon.)'') hints,
    (3) using a human-annotated $\gC_\text{task}$ (``Oracle (Anno.)'').
}
\label{fig:formalize_ablation}
\vspace{-3em}
\end{center}
\end{figure}

\subsubsection{Formalization Analysis}
\label{subsubsec:formalize_ablation}

We first investigate the necessity and impact of our core contribution, the $\gC_\text{task}$ constraint, by comparing our method against different reasoning strategies in~\Cref{fig:formalize_ablation}.
The results strongly confirm our central hypothesis.
Simply prompting the VLM to ``pay attention to the reference frame and objective in the query'' (``Tool (Prompt)'') only yields a negligible improvement on unconstrained tool integration.
This empirically suggests that the VLM's unconstrained planning process remains fundamentally flawed and unreliable, even when weakly guided by hints.
In comparison, the introduction of our formal $\gC_\text{task}$ constraint (``Ours'') delivers a substantial performance boost, far surpassing all unconstrained methods.
This demonstrates that a deterministic and verifiable constraint is essential for bridging the VLM's semantic-to-geometric gap, as it forces the VLM to first establish what to solve before determining how to solve it.
Furthermore, we explore the theoretical upper bound using a human-annotated oracle formalization (``Oracle (Anno.)'').
The gap between our method (47.6\%) and oracle (49.5\%) is relative small.
As revealed in~\Cref{subsec:error_ablation}, the $\gF_\text{formalize}$ stage achieves $\sim$70\% accuracy, confirming the formalization task is well within the VLM's capabilities.

\begin{figure}[t]
\begin{center}
\centerline{\includegraphics[width=0.95\linewidth]{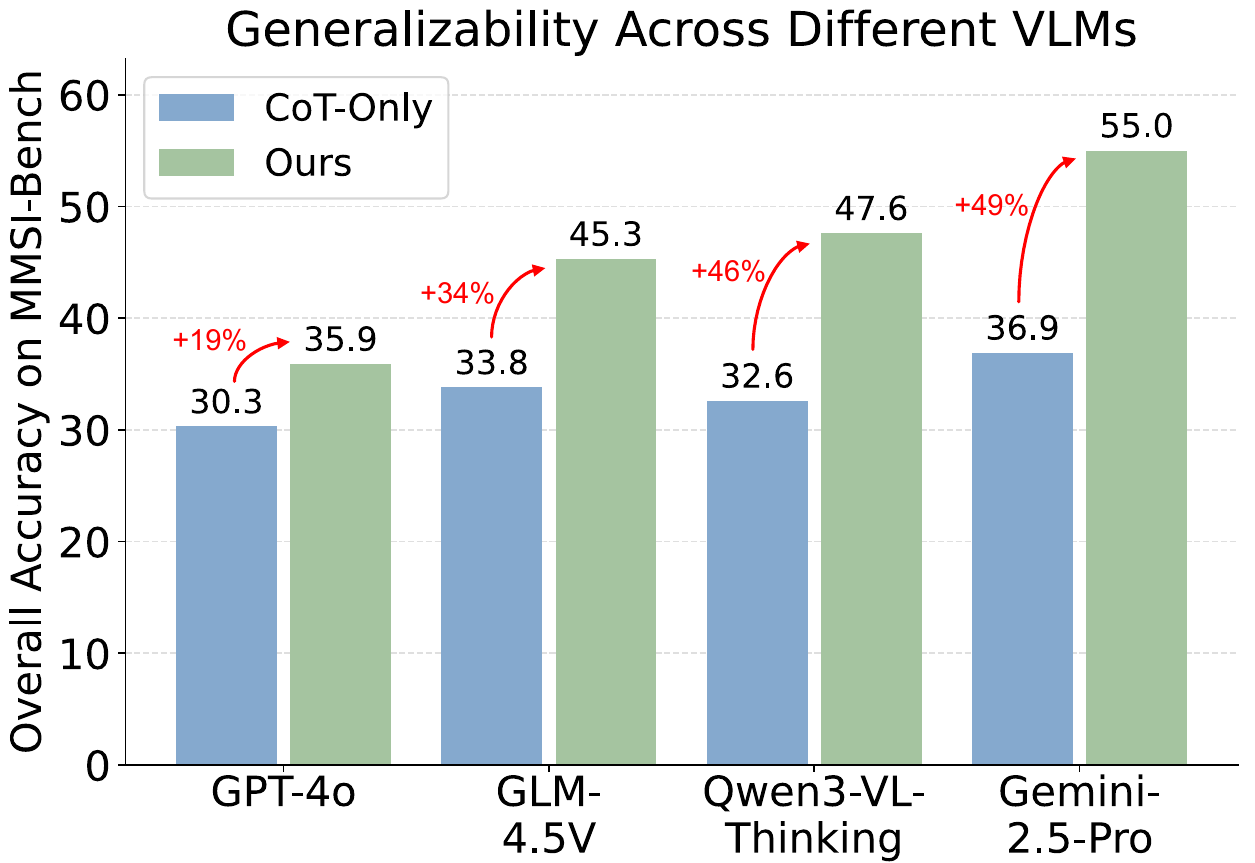}}
\caption{
    \textbf{Ablation Study on Generalizability across Different VLMs.}
    Our \METHOD{} achieves an average of 37\% relative performance improvement across all tested foundation VLMs.
}
\label{fig:generalization_ablation}
\vspace{-3em}
\end{center}
\end{figure}

\subsubsection{Generalizability Across VLMs}
\label{subsubsec:generalization_ablation}

We assess the generalizability of our \METHOD{} paradigm by applying it to several leading foundation VLMs, including GLM-4.5V~\cite{hong2025glm}, GPT-4o~\cite{hurst2024gpt}, and Gemini-2.5-Pro~\cite{comanici2025gemini}.
As shown in~\Cref{fig:generalization_ablation}, \METHOD{} proves to be a highly generalizable architectural solution, substantially enhancing the spatial reasoning capabilities of every VLM tested compared to their CoT-only baselines.
We observe that the magnitude of this enhancement appears to correlate strongly with the VLM's inherent agentic proficiency and their baseline spatial reasoning capability.
It is most evident that Gemini-2.5-Pro, which holds the strongest CoT-only baseline on MMSI-Bench (36.9\%), also achieves the most dramatic gain (+49\%), rising to 55.0\%.
On the other hand, the improvement on GPT-4o, while significant, is more modest (+19\%). 
We attribute it to its suboptimal agentic reasoning capability and coding skills.
Through introduction of formal task constraint $\gC_\text{task}$, our paradigm serves as a catalyst, successfully unlocking and guiding the VLM's powerful execution engine towards the robust spatial reasoning across a diverse set of SOTA models.

\begin{figure*}[t]
\centering
\includegraphics[height=12em]{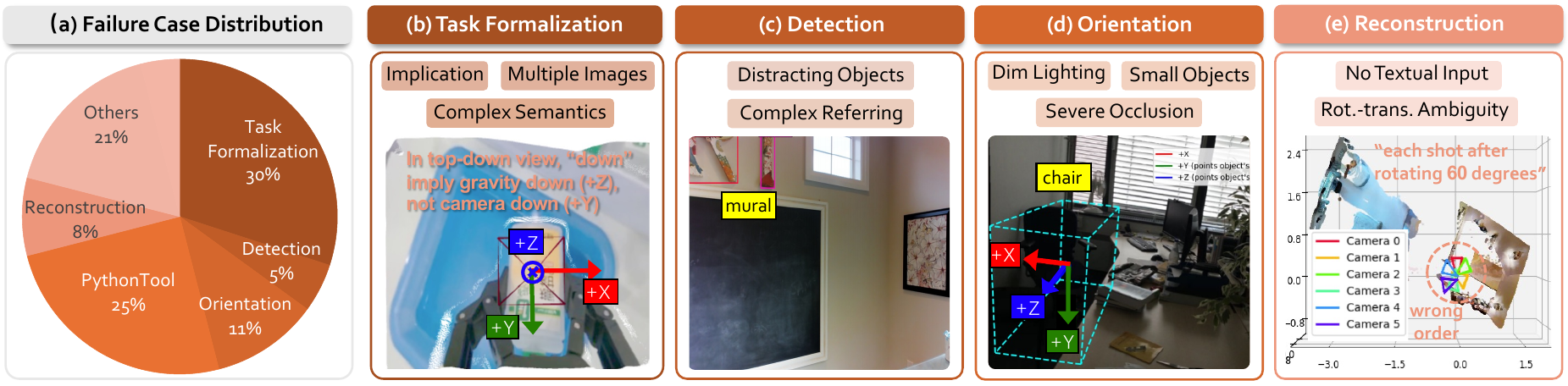}
\caption{
    \textbf{Error Attribution and Failure Cases.}
    We provide a detailed error attribution analysis to identify the main failure modes within the VLM's reasoning trajectory.
}
\label{fig:error_attr}
\end{figure*}

\subsubsection{Component Contribution}
\label{subsubsec:component_ablation}

We quantify the importance of each component in the \METHOD{}, as presented in~\Cref{tab:component_ablation}.
This analysis reveals improvements in two distinct parts.
First, building a standard tool-integrated agent by adding tool integration (+4.2 points), knowledge-augmented code generation (KACG, +1.9 points), and visual feedback (+1.4 points) provides a cumulative +7.5 points gain over the CoT-only baseline.
The second part, the introduction of $\gF_\text{formalize}$, brings an additional massive improvement, increasing the overall accuracy by +7.5 points.
This result strongly validates that constraining the VLM's planning via a formal $\gC_\text{task}$ is essential to prevent flawed reasoning within its lossy semantic space.

\begin{table}[t]
\caption{
    \textbf{
        Ablation Study on Each Component in \METHOD{}.
    }
    Here, ``KACG'' denotes applying knowledge-augmented code generation, and ``Feedback'' denotes applying the VLM to manage tool feedback and resolve ambiguity.
}
\vspace{-1em}
\label{tab:component_ablation}
\begin{center}
\resizebox{\linewidth}{!}{%
\begin{tabular}{cccc|c}
    \toprule
    Tool Integration & KACG & Feedback & $\gC_\text{task}$ & MMSI-Bench \\
    \midrule
    $\usym{2717}$ & $\usym{2717}$ & $\usym{2717}$ & $\usym{2717}$ & 32.6 \\
    \checkmark    & $\usym{2717}$ & $\usym{2717}$ & $\usym{2717}$ & 36.8 \\
    \checkmark    & \checkmark    & $\usym{2717}$ & $\usym{2717}$ & 38.7 \\
    \checkmark    & \checkmark    & \checkmark    & $\usym{2717}$ & 40.1 \\
    \midrule
    \rowcolor{gray!10} \checkmark    & \checkmark    & \checkmark    & \checkmark    & \textbf{47.6} \\
    \bottomrule
\end{tabular}
}
\end{center}
\vspace{-1em}
\end{table}

\subsection{Error Attribution and Failure Cases}
\label{subsec:error_ablation}

A key advantage of \METHOD{} paradigm is its verifiable and interpretable nature, which allows us to trace the reasoning pathway and perform detailed error attribution.
As shown in \Cref{fig:error_attr} (a), this analysis pinpoints the current bottlenecks, attributing failures to either the VLM's initial formalization or the subsequent tool orchestration.

\vspace{+0.5mm}
\noindent\textbf{Errors in $\gF_\text{formalize}$.}
Failures in the initial $\gF_\text{formalize}$ stage account for 30\% of all errors.
Given this is the first step of the paradigm, it indicates the VLM achieves approximately 70\% accuracy in correctly formalizing the task constraint $\gC_\text{task}$.
A deeper analysis reveals these failures primarily lie in challenging cases involving complex semantics, ambiguity in multiple images, or ignored implications.
For instance, as shown in~\Cref{fig:error_attr} (b), when asked about a top-down view, the VLM fails to grasp the query's implication that ``down'' referred to the direction of gravity, defaulting instead to ``camera down'' and establishing an incorrect reference frame.

\vspace{+0.5mm}
\noindent\textbf{Errors in $\gF_\text{compute}$.}
The remaining 70\% of errors occur during $\gF_\text{compute}$ stage.
Perception failures ($\sim$24\%) are a major bottleneck, particularly in ``Reconstruction'' and ``Orientation''.
A typical reconstruction failure, shown in~\Cref{fig:error_attr} (e), is caused by the inability of the underlying VGGT~\cite{wang2025vggt} to accept textual input.
The query's textual input, ``each shot after rotating 60 degrees'' provides a deterministic rotational sequence.
However, the VGGT model, which cannot accept this textual input, parameterize the scene incorrectly, resulting in the ``wrong order'' of cameras and a flawed geometric foundation.
Errors from ``Python Tool'' (25\%) are also significant, often stemming from forgotten coordinate transformations or lacking nuanced problem-solving logic, such as identifying a principal direction.
Besides, ``Other'' (21\%) errors capture issues like incorrect parameter passing between tools, exhausting the predefined budget (\eg{}, a maximum of 15 turns), and \etc{}

\section{Conclusion}

In this work, we introduce \METHOD{}, a training-free agentic paradigm designed to bridge the VLM's semantic-to-geometric gap in spatial reasoning.
We address it through leveraging a formal task constraint, transforming the ambiguous spatial query into a deterministic mathematic problem with constraints, preventing the VLM reasoning about the geometric details within its lossy semantic space.
As demonstrated experimentally, \METHOD{} establishes a new state-of-the-art on multiple challenging spatial reasoning benchmarks, showcasing a effective and generalizable pathway for robust spatial reasoning.

\vspace{+0.5mm}
\noindent\textbf{Limitations and Future Prospects.}
The \METHOD{} paradigm, involving iterative tool calls and VLM interactions, is computationally more costly than simple end-to-end CoT reasoning.
However, this trade off yields a more robust and verifiable reasoning pathway.
Furthermore, we believe the structured outputs from our $\gF_\text{formalize}$ and $\gF_\text{compute}$ stages can serve as a valuable source of supervision, such as a process reward in reinforcement learning, for training more efficient end-to-end spatial VLMs in the future.
Besides, our current toolbox is primarily designed for image-based spatial reasoning.
A key direction for future work is to extend this geometrically-constrained framework by incorporating tools for temporal reasoning and motion tracking, thereby addressing a broader range of spatial intelligence tasks.

{
    \small
    \bibliographystyle{ieeenat_fullname}
    \bibliography{main}
}

\appendix

\section{Spatial Task Constraint}
\label{sec:app_constraint}

As introduced in the main paper, the core of \Method{} (\METHOD{}) paradigm is the \textbf{formal task constraint} $\gC_\text{task}$.
It serves as a deterministic bridge to resolve the fundamental semantic-to-geometric gap, effectively decoupling the VLM's role into a semantic analyst and a constrained task solver.
Recall that $\gC_\text{task}$ is formally defined as a tuple containing two key sub-constraints: the \textbf{Reference Frame Constraint} ($\gC_\gR$), which defines the coordinate system, and the \textbf{Objective Constraint} ($\gC_\gO)$, which specifies what to measure within that frame.
We guide the Visual Language Model to automatically generate $\gC_\text{task}$ (for specific prompts used in \METHOD{}, refer to \Cref{sec:app_prompt}).
In this section, we first elaborate on the universality of this constraint across different spatial task domains.
%

\subsection{Universality of \texorpdfstring{$\gC_\text{task}$}{$C_\text{task}$} in Spatial Tasks}
\label{subsec:app_universal}

The $\gC_\text{task}$ is not a rigid definition fixed to a single problem type, but rather a general principle for a wide range of spatial tasks with semantic-to-geometric gap.
The core idea is to leverage the VLM's semantic advantage to formalize the most significant geometric ambiguity of the task.
The nature of this ambiguity shifts depending on the task domain.

\vspace{+0.5mm}
\noindent\textbf{Spatial Understanding and Reasoning.}
For spatial reasoning tasks, such as those evaluated in the main paper~\cite{yang2025mmsi,yang2025thinking,jia2025omnispatial,yin2025spatial,li2025spatialladder}, the objective  is typically simple and explicitly stated in the query (\eg{}, ``what is the relative position?'' or ``which object is wider?'').
Therefore, the objective constraint $\gC_\gO$ is often trivial to formalize.
The primary geometric ambiguity lies in the reference frame constraint $\gC_\gR$.
A query like ``where is the table relative to you?'' is unsolvable until the reference frame (``you'') is geometrically grounded.
\METHOD{}'s formalization of $\gC_\gR$ (\eg{}, object-based, camera-based, direction-based frames)  is specifically designed to resolve this ambiguity.

\vspace{+0.5mm}
\noindent\textbf{Robotic Manipulation and Interaction.}
Conversely, for robotic manipulation and interaction tasks~\cite{yang2025embodiedbench,zhang2025vlabench,li2024embodied,lu2025bench,choi2024lota}, the reference frame constraint is often simple.
The frame is typically the robot's egocentric perspective or a fixed world frame aligned with the primary camera.
The geometric ambiguity shifts entirely to the objective constraint $\gC_\gO$.
A command like ``pour tea into the cup'' has a trivial $\gC_\gR$ but a highly complex $\gC_\gO$.
The objective is not a simple measurement but a complex, multi-stage procedure involving affordances, contact points, and trajectories.
For example, ReKep~\cite{huang2025rekep} tackles the manipulation by instructing the VLM to formalize the objective $\gC_\gO$ as a set of \textit{Relational Keypoint Constraints}.
These constraints $\gC_\gO$ are literally generated as cost functions written by VLM in Python that map 3D keypoints to a numerical cost.
The cost functions are then passed to an inverse kinematics solver ($\gF_\text{compute}$) to find the optimal robot action.
Similarly, EmbodiedCoder~\cite{lin2025embodiedcoder} formalizes the $\gC_\gO$ as an executable program.
The VLM is prompted to first generate code for \textit{geometric parameterization}, fitting point clouds to functional primitives like a rectangle and a hinge axis for a door.
Through generating a precise, parametric motion that conforms to the geometric shape just defined, the Python interpreter ($\gF_\text{compute}$) simply executes the code to produce the final waypoints.

These works are not in conflict with our proposed task constraint $\gC_\text{task}$.
Instead, they can serve as complementary examples of its core principle.
They demonstrate how the $\gC_\gO$ for complex manipulation and interaction can be formalized as code, cost functions, or geometric constraints.
These constraints are then passed to a solver, just as \METHOD{} proposes.
This confirms the universality of the $\gC_\text{task}$:
whether the ambiguity lies in the reference frame or the objective, the first and most critical step is to use the VLM's semantic strength to formalize a deterministic, geometrically-sound constraint to bridge the semantic-to-geometric gap.

\subsection{Generalizability of \texorpdfstring{$\gR$}{$R$} Definintion}
\label{subsec:app_generalize_reference}

\begin{table}[t]
\caption{
    \textbf{
        Ablation Study on Task Constraint.
    }
}
\vspace{-1.5em}
\label{tab:app_task_constraint}
\begin{center}
\resizebox{0.92\linewidth}{!}{%
\begin{tabular}{ccc|c}
    \toprule
    Tool Integration & Ref. $\gC_\gR$ & Obj. $\gC_\gO$ & MMSI-Bench \\
    \midrule
    \rowcolor{gray!10} \checkmark    & \checkmark    & \checkmark    & \textbf{47.6} \\
    \midrule
    \checkmark    & \checkmark    & $\usym{2717}$ & 46.4 \\
    \checkmark    & $\usym{2717}$ & \checkmark    & 41.0 \\
    \checkmark    & $\usym{2717}$ & $\usym{2717}$ & 40.1 \\
    \midrule
    $\usym{2717}$ & \checkmark    & \checkmark    & 33.5 \\
    \bottomrule
\end{tabular}
}
\end{center}
\vspace{-1em}
\end{table}

We define three types of reference frame $\gR$ in \METHOD{}: object-based, camera-based and direction-based reference frame, providing a robust and flexible framework.
We find that these categories are sufficient to cover the vast majority of static spatial reasoning queries encountered in existing benchmarks~\cite{yang2025mmsi, jia2025omnispatial, li2025spatialladder, yin2025spatial}.
As spatial reasoning advances toward more complex, dynamic, and abstract scenarios, we identify key limitations and challenges for the current implementation of $\gC_\gR$. 
These represent important avenues for future research.

\vspace{+0.5mm}
\noindent\textbf{Dynamic and Time-Varying Reference Frame.}
A significant challenge arises in video-based spatial reasoning~\cite{yang2025thinking}, particularly in tasks involving continuous navigation or long-horizon agent actions.
For example, an instruction like, ``Move forward to the right first, then move backward to the right, and finally turn left,'' involves a $\gC_\gR$ that is continuously changing at each step, contingent on the agent's previous state.
Solving this requires extending the $\gC_\gR$ formalization to become a time-dependent function $\gC_\gR(t)$, capable of tracking and updating the agent's pose and orientation throughout a sequence.

\vspace{+0.5mm}
\noindent\textbf{Frames from Abstract Concepts.}
This challenge emerges when the reference entity is not a rigid, easily-definable object.
A query like ``the living room is south of the kitchen'' poses a significant problem. 
It is often impossible to compute a meaningful direction vector between the geometric centroids of two abstract areas or regions.
Currently, \METHOD{} relies on proxies, for example, if such direction (from kitchen to living room) aligns with the camera's view from the background to the foreground, we might substitute $-Z_\text{cam}$ as the direction vector.
This workaround, however, can introduce cumulative errors and lacks generalizability.
Future work could explore novel methods to address this.
One promising direction is to empower the VLM to directly annotate the reference frame, \ie{}, outputting two points in the image whose corresponding 3D vector defines the abstract direction.

\section{More Ablation Studies}
\label{subsec:app_more_ablation}

To further explore the proposed formal task constraint $\gC_\text{task}$ and the stability of \METHOD{}, we present three additional ablation studies.
These experiments are designed to 
(1) precisely quantify the relative importance of the reference frame constraint ($\gC_\gR$) versus the objective constraint ($\gC_\gO$) for spatial reasoning tasks, 
(2) validate the constraint's effectiveness in improving the VLM's internal, tool-free reasoning, and
(3) evaluate the stability and robustness of the \METHOD{} paradigm.

\vspace{+0.5mm}
\noindent\textbf{Importance of $\gC_\gR$ and $\gC_\gO$ in Spatial Reasoning.}
To validate our claim in \Cref{subsec:app_universal} that $\gC_\gR$ represents the primary geometric ambiguity in spatial reasoning tasks, we conduct a detailed ablation on the sub-components of $\gC_\text{task}$.
As shown in~\Cref{tab:app_task_constraint}, removing the objective constraint ($\gC_\gO$) results in a minor 1.2 point performance drop.
This demonstrates that for spatial reasoning queries, the objective is often simple and clearly stated, allowing the task solver to infer it from the query during the $\gF_\text{compute}$ stage.
In contrast, removing the reference frame constraint ($\gC_\gR$) causes a 6.6 point performance drop.
This suggests that $\gC_\gR$ is the most critical component in spatial reasoning, as it resolves the core geometric ambiguity of ``from where'' that VLMs cannot solve in their lossy semantic space.

\vspace{+0.5mm}
\noindent\textbf{$\gC_\text{task}$ without Tool Integration.}
The $\gC_\text{task}$ is not a prompt that can fix the VLM's internal spatial reasoning.
Instead, it unlocks the VLM's agentic reasoning capability and coding skills by constraining the subsequent computation stage ($\gF_\text{compute}$).
This is confirmed by the results in~\Cref{tab:app_task_constraint}, where we compare \METHOD{} with a VLM that receives $\gC_\text{task}$ as the prompt but relies solely on chain-of-thought (CoT) reasoning.
With the constraint as a textual hint, it only yields a negligible 0.9 point improvement.
Even when told the reference frame and objective, the VLM still cannot bypass the internal flawed spatial imagination and high-precision computation in its lossy semantic space.

\vspace{+0.5mm}
\noindent\textbf{Stability and Robustness Analysis.}
A key consideration for any agentic framework, especially one involving multiple VLM calls and tool interactions, is the stability of its results. 
The probabilistic nature of VLMs could potentially lead to high variance in final performance.
To assess the robustness of \METHOD{}, we conduct $N=10$ independent evaluation runs on the complete MMSI-Bench dataset, using the same Qwen3-VL-Thinking~\cite{qwen2025qwen3vl} for each run. 
All settings are kept identical as in the Table 1 (main paper).
The mean accuracy and the standard deviation across all 10 runs is 47.6 $\pm$ 0.3.
The results demonstrate a very low standard deviation, indicating that the \METHOD{} framework is highly stable.
This stability is a direct benefit of our core design: by forcing the VLM to first generate a deterministic formal task constraint $\gC_\text{task}$, we significantly reduce the stochasticity and ambiguity in the subsequent $\gF_\text{compute}$ stage. 
The task solver operates within the non-negotiable geometric bounds defined by $\gC_\text{task}$, leading to a consistent and verifiable reasoning pathway.

\section{More Implementation Details}
\label{sec:app_more_impl}

\subsection{Visual Foundation Models}
\label{subsec:app_vfm}

We deploy several Visual Foundation Models (VFMs) for agent to parameterize the visual world, facilitating the deterministic $\gF_\text{compute}$ stage constrained by $\gC_\text{task}$.
\begin{itemize}
    \vspace{+0.5mm}
    \item \textbf{VGGT}~\cite{wang2025vggt}.
    Visual Geometry Grounded Transformer (VGGT) is a large feed-forward model that infers key 3D attributes from one or multiple images. 
    It predicts camera parameters, point maps, and depth maps for all input views.
    Within \METHOD{}, VGGT serves as the primary geometry parameterization engine for 3D reconstruction.
    \vspace{+0.5mm}
    \item \textbf{MoGe-2}~\cite{wang2025moge2}.
    The Monocular Geometry (MoGe) estimation model is designed to recover 3D point maps with metric scale from a single image. 
    It achieves this by decoupling the problem, predicting both an affine-invariant point map and a separate global scale factor. 
    \METHOD{} leverages this unique capability to derive the correct real-world scale for the scene.
    \vspace{+0.5mm}
    \item \textbf{GroundingDINO}~\cite{liu2024grounding}.
    GroundingDINO is an open-set object detector that combines a transformer-based detector with grounded language pre-training. 
    This architecture enables it to detect arbitrary objects specified by natural language, such as category names or referring expressions. 
    It serves as a specialized detection tool in \METHOD{}.
    \vspace{+0.5mm}
    \item \textbf{SAM-2}~\cite{ravi2024sam}.
    SAM-2 is a foundation model for promptable visual segmentation in both images and videos.
    It generalizes the original SAM by incorporating a streaming memory architecture to handle temporal data. 
    \METHOD{} uses SAM-2 as a bridge connecting pixels and boxes, allowing us to extract object point clouds from the VG-GT output based on the boxes.
    \vspace{+0.5mm}
    \item \textbf{Orient Anything}~\cite{wang2024orient}.
    Orient Anything is trained on rendered 3D models to estimate the 3D orientation of an object from a single, free-view image. 
    It predicts the object's azimuth, polar, and rotation angles relative to the camera. 
    This capability is crucial for \METHOD{} to construct a object-based reference frame.
\end{itemize}
\vspace{+0.5mm}
Note that these foundation models are not provided directly to the agent. 
Instead, we wrap them and offer some abstract tool interfaces as APIs for invocation (see~\Cref{subsec:app_tool_interfaces}).

\subsection{Tool Interfaces}
\label{subsec:app_tool_interfaces}

The GCA agent's $\gF_\text{compute}$ stage is driven by a discrete set of 8 exposed tool APIs. 
These APIs form the agent's action space, encapsulating the underlying VFMs.
\begin{itemize}
    \vspace{+0.5mm}
    \item \texttt{reconstruct}. 
    It ingests one or more images and produces a comprehensive 3D reconstruction. 
    Internally, it leverages VGGT and automatically selects the optimal reconstruction strategy. 
    If multiple, non-static images are provided, it first consults the VLM to identify common static objects for alignment. 
    The output includes the 3D world points, camera extrinsics, and intrinsics.
    \vspace{+0.5mm}
    \item \texttt{detect}. 
    It detects target objects in a single image based on a text prompt. 
    For capable VLMs like Qwen3-VL-Thinking~\cite{qwen2025qwen3vl}, we directly instruct the VLM itself to locate the target object through prompts.
    Otherwise, we use GroundingDINO as the detector.
    It returns the bounding boxes and corresponding labels.
    \vspace{+0.5mm}
    \item \texttt{project\_box\_to\_3d\_points}. 
    It takes a 2D bounding box and projects it into the 3D world coordinate system defined by the VGGT model output.
    Internally, it first uses SAM-2 to convert the bounding box into a precise pixel mask, then filters the points using this mask.
    \vspace{+0.5mm}
    \item \texttt{predict\_obj\_pose}.
    It computes the 6-DoF semantic pose of an object, which is essential for establishing an object-based reference frame. 
    This tool first calls \texttt{project\_box\_to\_3d\_points} to find the object's 3D centroid, and then calls Orient Anything to determine its 3D orientation. 
    It then combines these to return the final object-to-world transformation matrix.
    \vspace{+0.5mm}
    \item \texttt{estimate\_scale}.
    This tool is called when metric measurements (\eg{}, ``meters'', ``feet'') are required. 
    It aligns MoGe-2's metric depth with the VGGT model's relative depth prediction to compute a single scale factor that converts the entire reconstruction into meters.
    \vspace{+0.5mm}
    \item \texttt{ocr}.
    It performs optical character recognition (OCR) on an image using the EasyOCR library. 
    It returns a list of recognized texts and their bounding boxes.
    \vspace{+0.5mm}
    \item \texttt{analyze\_motion}.
    It analyzes pixel-level motion between two sequential images using a Farneback optical flow algorithm~\cite{farneback2003two}.
    It is used to infer subtle camera movements that may be too small for full 3D reconstruction.
    \vspace{+0.5mm}
    \item \texttt{code}.
    This is the agent's primary computation engine. It generates and executes Python code within a sandbox environment. 
    It tasks a set of context variables (\eg{}, poses, points) and an natural language description (\eg{}, request and description of the variables) as input.
    The code is generated using a knowledge-augmented strategy, where relevant geometric formulas are injected into the prompt, ensuring the computation is both deterministic and sound.
\end{itemize}

\subsection{Agentic Framework}
\label{subsec:app_framework}

The \METHOD{} paradigm is implemented as a high-throughput, modular system. 
The core VLM deployment and the perception tool suite are physically decoupled to ensure scalability and robustness.

\vspace{+0.5mm}
\noindent\textbf{System Backend and State Management.} 
The entire system is built using Ray~\cite{moritz2018ray} and LangGraph.
LangGraph is used to define and manage the agent's state and orchestrate the two-stage, graph-based reasoning flow (\ie{}, $\gF_\text{formalize}$ followed by the $\gF_\text{compute}$ loop).

\vspace{+0.5mm}
\noindent\textbf{Tool Suite and VFMs Deployment.} 
The perceptual and computational tools (listed in \Cref{subsec:app_tool_interfaces}) are encapsulated as independent Ray Serve actors. 
This microservice architecture allows \METHOD{} to make concurrent perception requests (\eg{}, running reconstruct and detect in parallel), enabling high parallelism and automatic scaling. 
This entire tool suite is deployed on 2 NVIDIA A100 GPUs.

\vspace{+0.5mm}
\noindent\textbf{VLM Roles and Deployment.}
In \METHOD{}, a single VLM fulfills the three distinct roles within the \METHOD{} paradigm:
\begin{itemize}
    \vspace{+0.5mm}
    \item \textbf{Semantic Analyst.}
    In the $\gF_\text{formalize}$ stage, it interprets the query and visual context to generate the formal $\gC_\text{task}$.
    \vspace{+0.5mm}
    \item \textbf{Tool Orchestrator.}
    In the $\gF_\text{compute}$ stage, it manages the ReAct-style tool call loop, resolves ambiguities, and generates natural language descriptions for the coder.
    \vspace{+0.5mm}
    \item \textbf{Coder.}
    It generates Python code for the \texttt{code} tool.
\end{itemize}
\vspace{+0.5mm}
The VLM deployment is separate from the tool suite.
For open-source models (\eg{}, Qwen3-VL-Thinking~\cite{qwen2025qwen3vl}), we use vLLM~\cite{kwon2023efficient} for efficient, high-throughput inference on 8 NVIDIA A100 GPUs.
For closed-source models (\eg{}, Gemini-2.5-Pro~\cite{comanici2025gemini}), we access them via their standard commercial APIs.
We employ different sampling parameters based on the VLM's role. 
For the semantic analyst and tool orchestrator roles, which require reasoning and flexibility, we use \texttt{TEMPERATURE=0.6} and \texttt{TOP\_P=0.95}. 
For the coder role, which demands deterministic and reliable output, we set \texttt{TEMPERATURE=0.0}. 
All roles use \texttt{MAX\_TOKENS=32768}.

\section{Evaluation Benchmark Details}
\label{sec:app_benchmark}

We evaluate \METHOD{} on multiple challenging spatial reasoning benchmarks. 
This section provides a detailed description of each benchmark and its subcategories, corresponding to the results presented in Table 1 of the main paper.

\subsection{MMSI-Bench}
\label{subsec:app_mmsi}

MMSI-Bench~\cite{yang2025mmsi} is a comprehensive benchmark designed to evaluate a VLM's ability to perform spatial reasoning by integrating information from multiple, distinct images. 
It is organized into four distinct subcategories:
\begin{itemize}
    \vspace{+0.5mm}
    \item \textbf{PR. (Positional Relationship).}
    This subcategory evaluates the model's ability to understand the relative positions between different objects, cameras and semantic regions (\eg{}, a kitchen) across multiple views.
    \vspace{+0.5mm}
    \item \textbf{Attr. (Attribute).} 
    This subcategory evaluates the model's ability to identify object attributes related to spatial properties, such as geometric properties (\eg{}, size, length) or visual characteristics (\eg{}, shape).
    \vspace{+0.5mm}
    \item \textbf{Mot. (Motion).} 
    This subcategory evaluates the model's ability to understand object or camera's movement.
    \vspace{+0.5mm}
    \item \textbf{MSR (Multi-Step Reasoning).}
    This subcategory evaluates the model's ability to perform complex reasoning by chaining multiple spatial understandings described above together to arrive at a final answer.
\end{itemize}

\subsection{MindCube-tiny} \label{subsec:app_mindcube}

MindCube-tiny is a subset of the MindCube benchmark~\cite{yin2025spatial}, which is designed to test Spatial Mental Modeling (SMM). 
The core task evaluates a VLM's ability to construct and manipulate a 3D mental model of a scene using only a limited set of 2D images as input. 
The ``tiny'' version is a smaller-scale version of the full benchmark. We evaluate on its three primary sub-tasks:
\begin{itemize}
    \vspace{+0.5mm}
    \item \textbf{Rot. (Rotation).}
    This task requires the model to infer the complete environment based on partial visual information, testing its understanding of sequential views and consistent spatial cues between images (\eg{}, lighting).
    \vspace{+0.5mm}
    \item \textbf{Ard. (Around).}
    This task requires the model to infer the scene from a novel viewpoint, testing its ability to interpolate and extrapolate its mental 3D model.
    \vspace{+0.5mm}
    \item \textbf{Amg. (Among).}
    This task requires the model to infer the 3D spatial arrangement based on four orthogonal views characterized by significant occlusion, testing its ability to establish consistency relationships across perspectives and reason about the relative positions of unseen objects.
\end{itemize}

\subsection{OmniSpatial} \label{subsec:app_omnispatial}

OmniSpatial~\cite{jia2025omnispatial} is a comprehensive benchmark designed to evaluate a broad spectrum of visual-based spatial intelligence capabilities in VLMs. 
The full benchmark consists of four categories. 
We primarily focus on the two subcategories most closely related to geometric perception:
\begin{itemize} 
    \vspace{+0.5mm} 
    \item \textbf{Pers. (Perspective Taking).}
    This subcategory assesses the model's understanding of 3D spatial relationships by adopting varied viewpoints, \eg{}, egocentric, allocentric, and hypothetical perspective.
    \vspace{+0.5mm} 
    \item \textbf{Dyn. (Dynamic Reasoning).}
    This subcategory assesses the model's understanding of object motion and judgments in uncertain or rapidly changing environments.
\end{itemize}
\vspace{+0.5mm}
We exclude the other two subcategories: ``Spatial Interaction'' (which focuses on diagrams and user-interface, \eg{}, terrain map) and ``Complex Logic'' (which involves abstract spatial reasoning, \eg{}, puzzles), as they are more centered on abstract or symbolic reasoning rather than the high-fidelity geometric perception that \METHOD{} is designed to solve.

\subsection{SPBench}
\label{subsec:app_spbench}

SPBench~\cite{li2025spatialladder} is a benchmark designed to evaluate both VLM's spatial reasoning in single view and multiple views:
\begin{itemize} 
    \vspace{+0.5mm} 
    \item \textbf{SI (Single Image).}
    This subset contains questions that test the model's understanding and reasoning capabilities within a single image, including absolute distance, object size, relative distance, and relative direction.
    \vspace{+0.5mm} 
    \item \textbf{MV (Multiple Views).}
    This subset requires the model to integrate information from multiple viewpoints to answer questions about relative position and object counts within a overlapping scene.
\end{itemize}

\subsection{CV-Bench}
\label{subsec:app_cvbench}

CV-Bench is a visual-centric benchmark that evaluates the spatial understanding capabilities of VLMs.
It is broadly composed of 2D and 3D reasoning tasks:
\begin{itemize} 
    \vspace{+0.5mm} 
    \item \textbf{2D (2D Relationship).} 
    This subcategory tests fundamental 2D spatial understanding, including 2D positional relationships and object counting. 
    \vspace{+0.5mm} 
    \item \textbf{3D (3D Relationship).} 
    This subcategory assesses the model's grasp of 3D concepts, such as depth analysis and 3D distance comparisons between objects in the scene.
\end{itemize}

\begin{figure}[t]
\begin{center}
\centerline{\includegraphics[width=\linewidth]{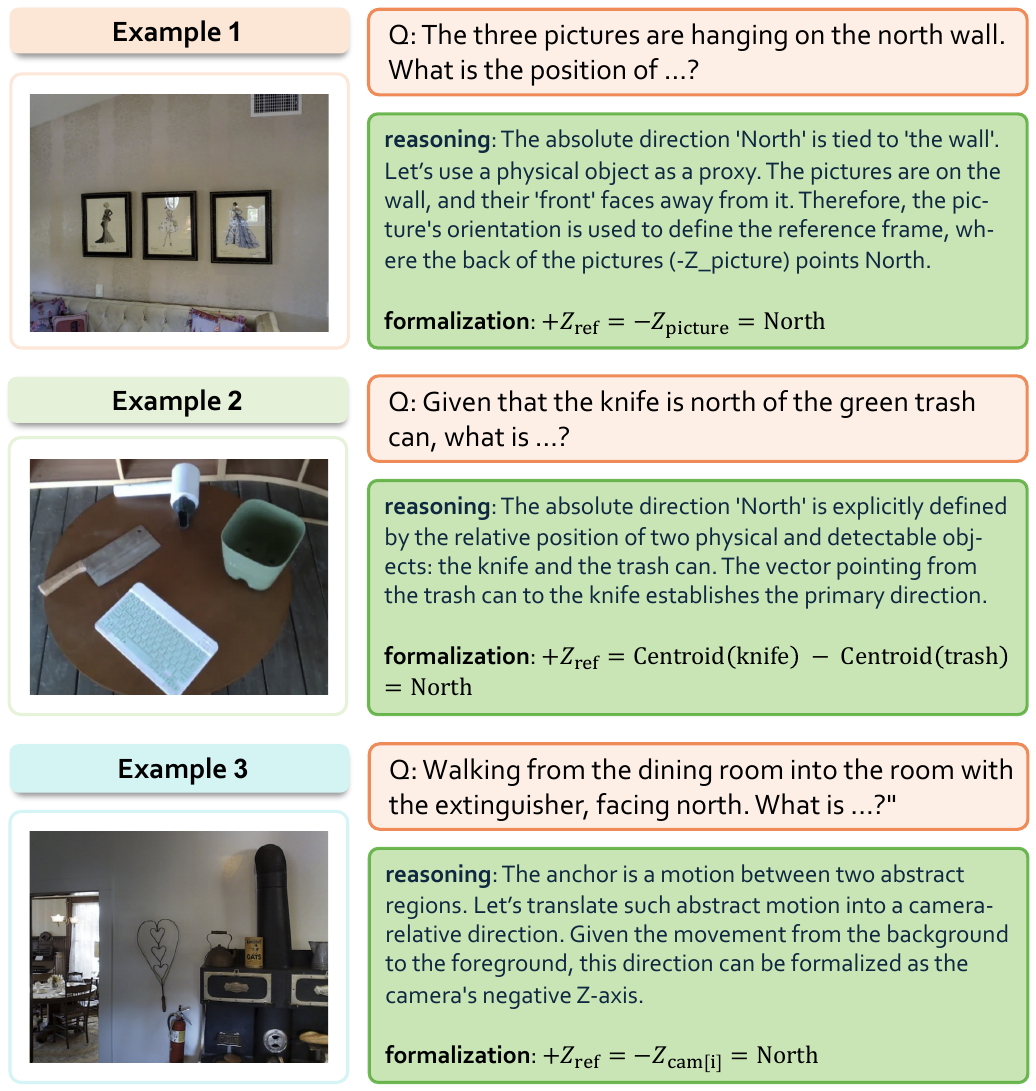}}
\caption{
    \textbf{In Context Examples Used in Formalizing Reference Frame.}
    The output format follows the prompt in~\Cref{tab:app_prompt_formalize_reference}.
}
\label{fig:app_in_context}
\vspace{-2.5em}
\end{center}
\end{figure}

\section{Prompts Used in \METHOD{}}
\label{sec:app_prompt}

We provide detailed prompts used in \METHOD{}, including task formalization (\Cref{tab:app_prompt_formalize_reference} and~\ref{tab:app_prompt_formalize_objective}), tool orchestration (\Cref{tab:app_prompt_tool}) and knowledge-augmented code generation (\Cref{tab:app_prompt_code} and \ref{tab:app_prompt_kacg}).
Besides, we also provide the in context examples used in the reference frame formalization (see~\Cref{fig:app_in_context} and~\Cref{tab:app_prompt_formalize_reference}).
\section{Qualitative Case Study}
\label{sec:app_case_study}

\begin{figure*}[t]
\begin{center}
\centerline{\includegraphics[width=0.79\linewidth]{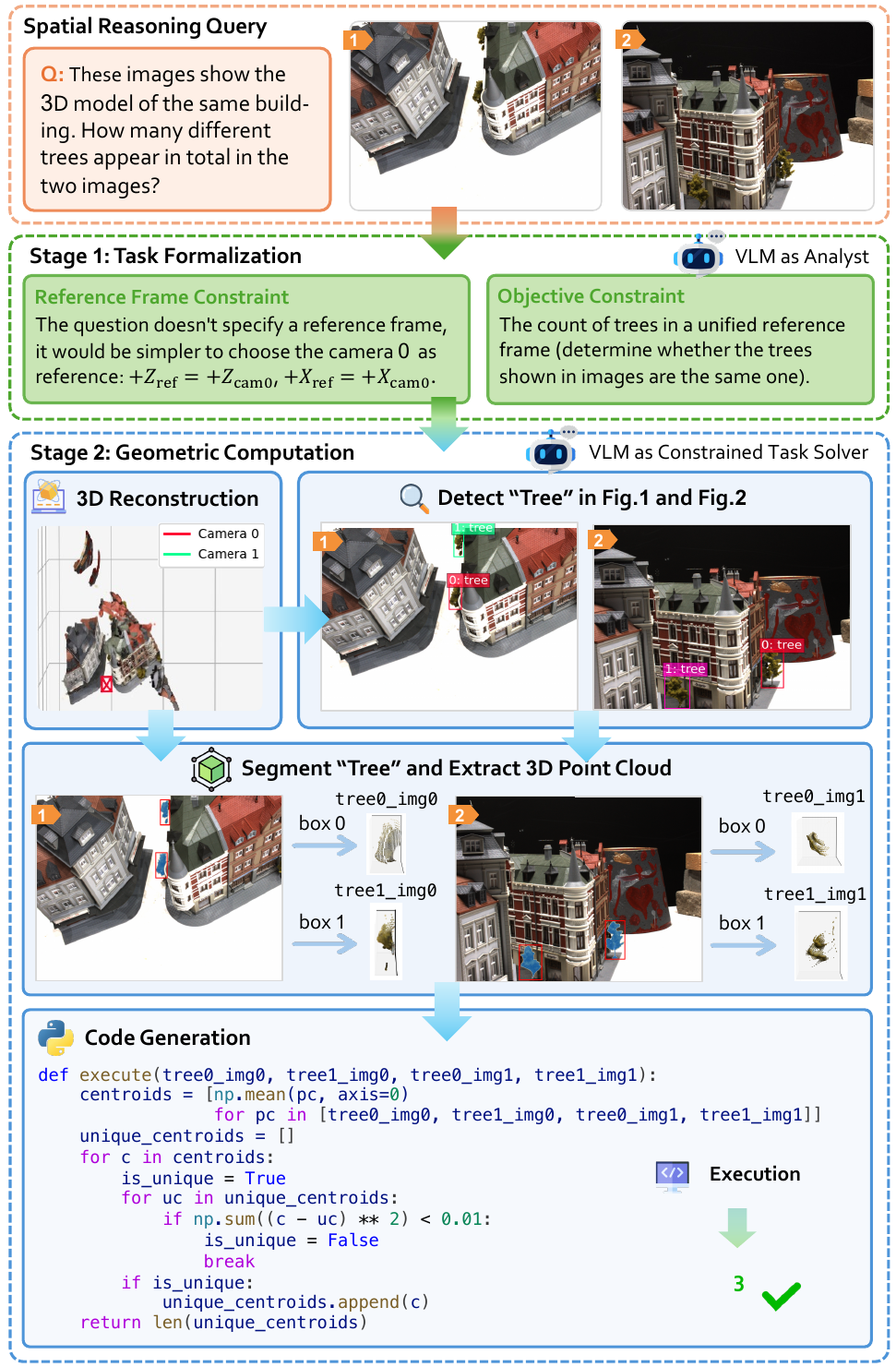}}
\caption{
    \textbf{Case Study \#1.}
    Unique object counting across multiple views.
}
\label{fig:app_case1}
\vspace{-2em}
\end{center}
\end{figure*}

\begin{figure*}[t]
\begin{center}
\centerline{\includegraphics[width=0.79\linewidth]{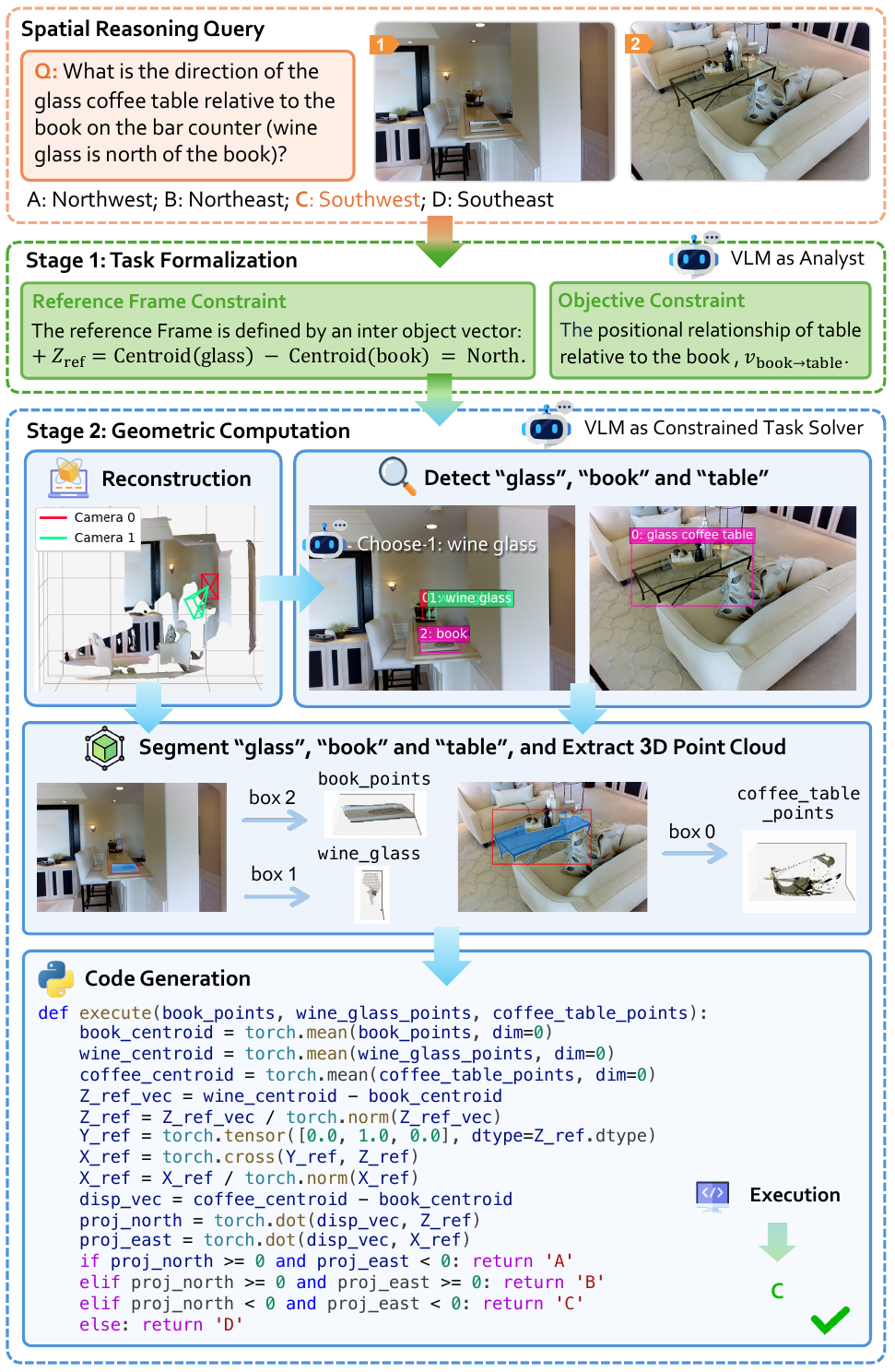}}
\caption{
    \textbf{Case Study \#2.}
    Direction-based reference frame.
}
\label{fig:app_case2}
\vspace{-2em}
\end{center}
\end{figure*}

\begin{figure*}[t]
\begin{center}
\centerline{\includegraphics[width=0.79\linewidth]{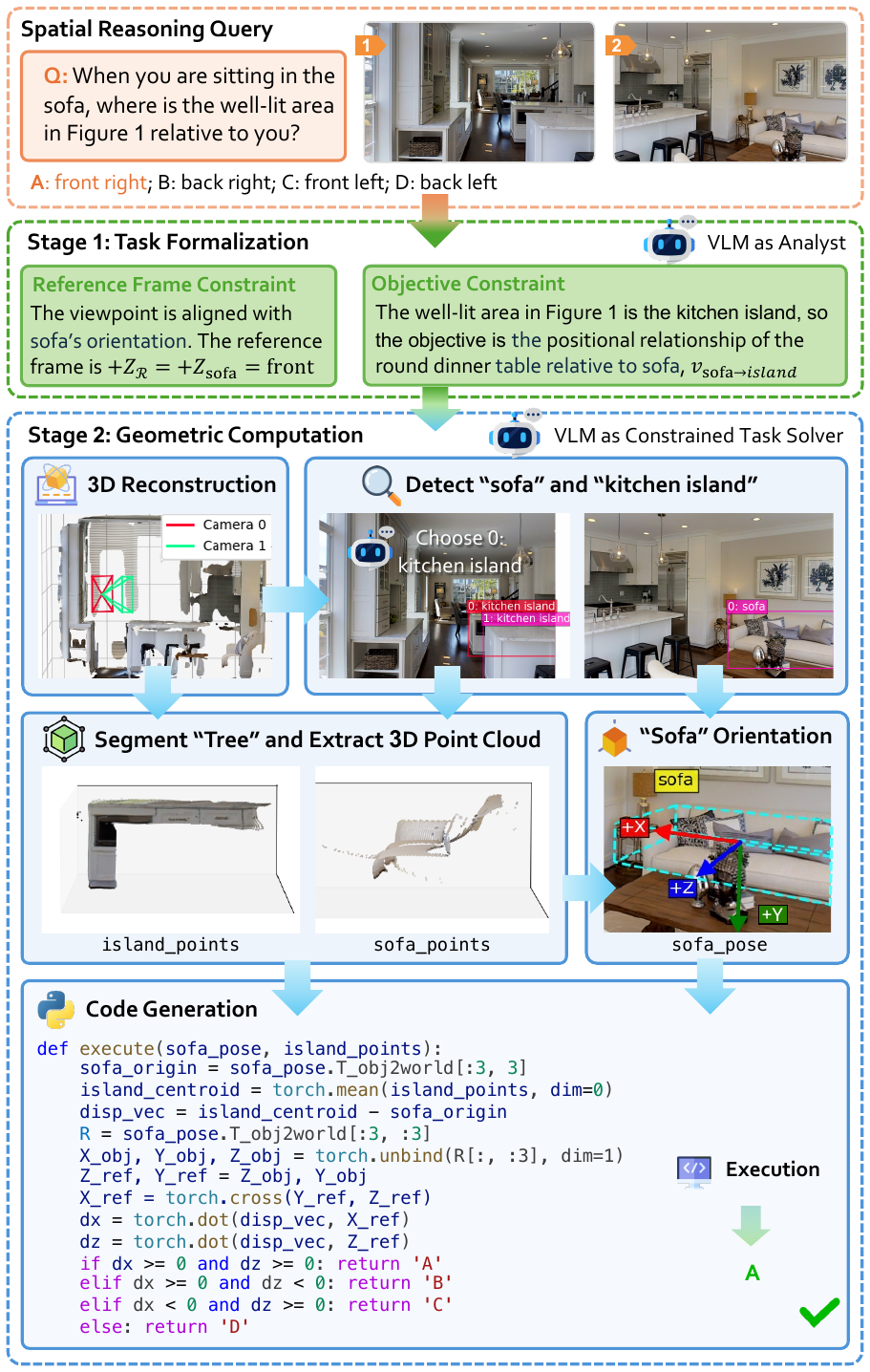}}
\caption{
    \textbf{Case Study \#3.}
    Object-based reference frame
}
\label{fig:app_case3}
\vspace{-2em}
\end{center}
\end{figure*}

\begin{figure*}[t]
\begin{center}
\centerline{\includegraphics[width=0.86\linewidth]{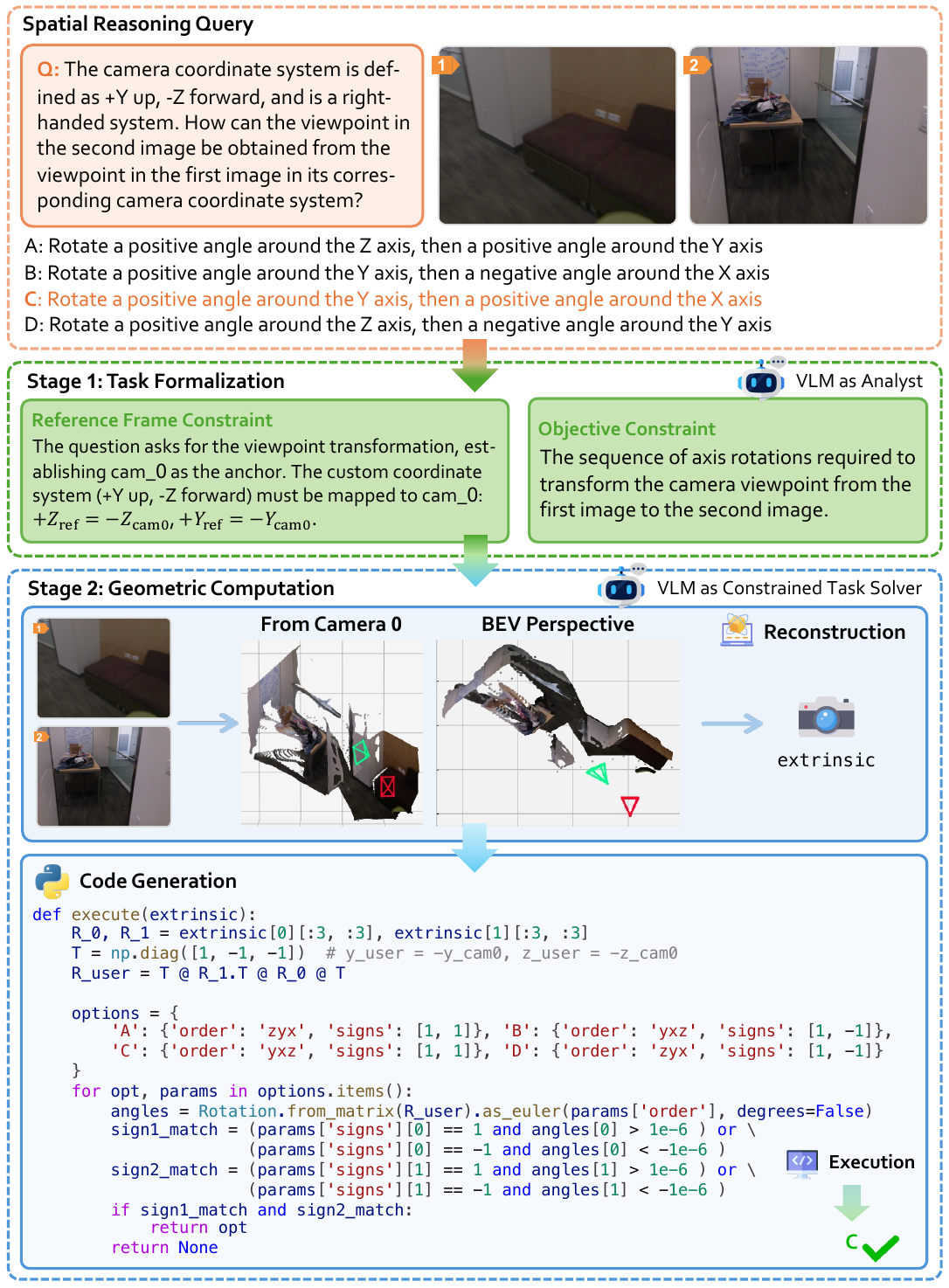}}
\caption{
    \textbf{Case Study \#4.}
    Camera rotation analysis.
}
\label{fig:app_case4}
\vspace{-2em}
\end{center}
\end{figure*}

\begin{figure*}[t]
\begin{center}
\centerline{\includegraphics[width=0.79\linewidth]{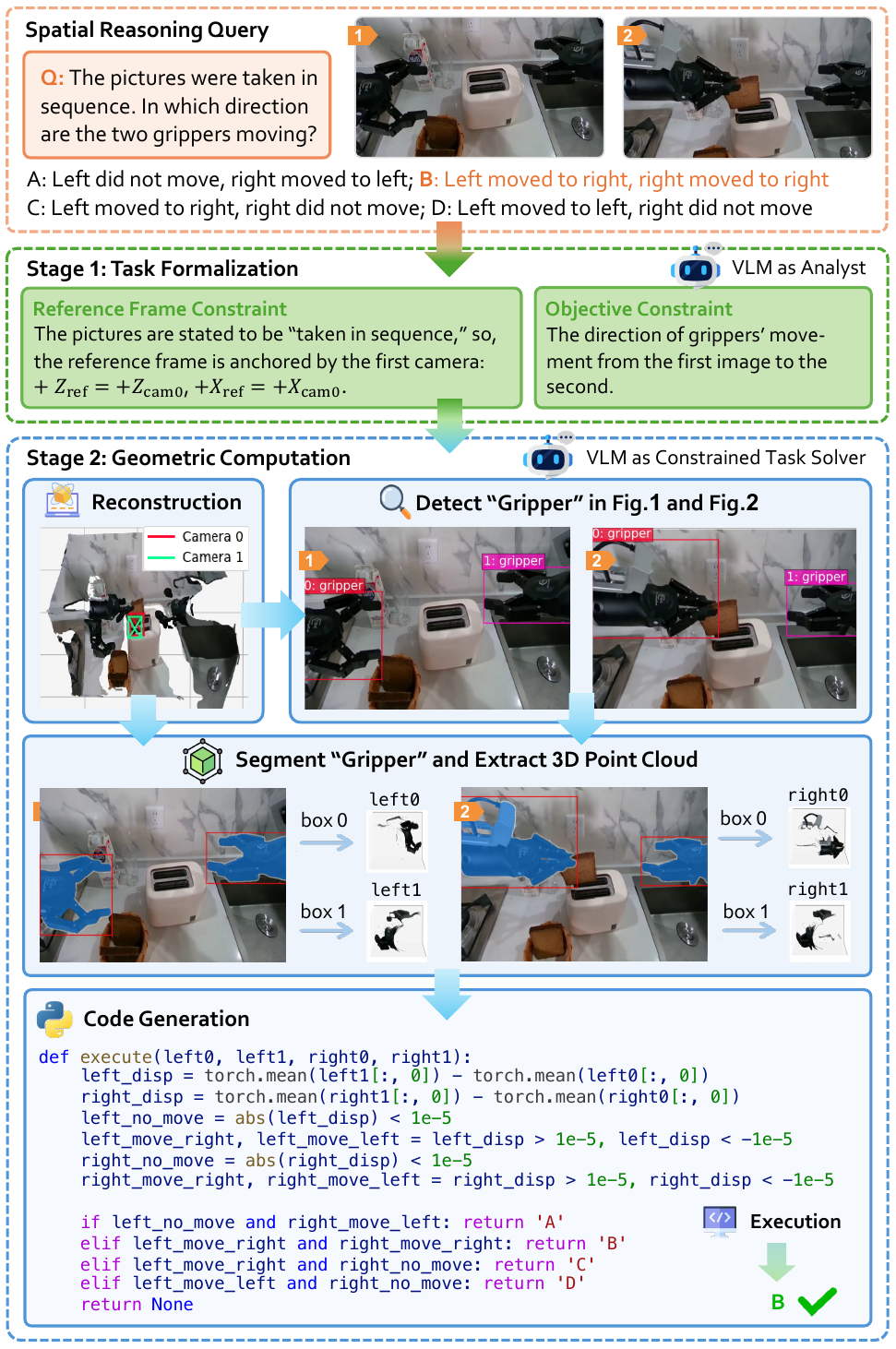}}
\caption{
    \textbf{Case Study \#5.}
    Object movement analysis.
}
\label{fig:app_case5}
\vspace{-2em}
\end{center}
\end{figure*}

\begin{figure*}[t]
\begin{center}
\centerline{\includegraphics[width=0.73\linewidth]{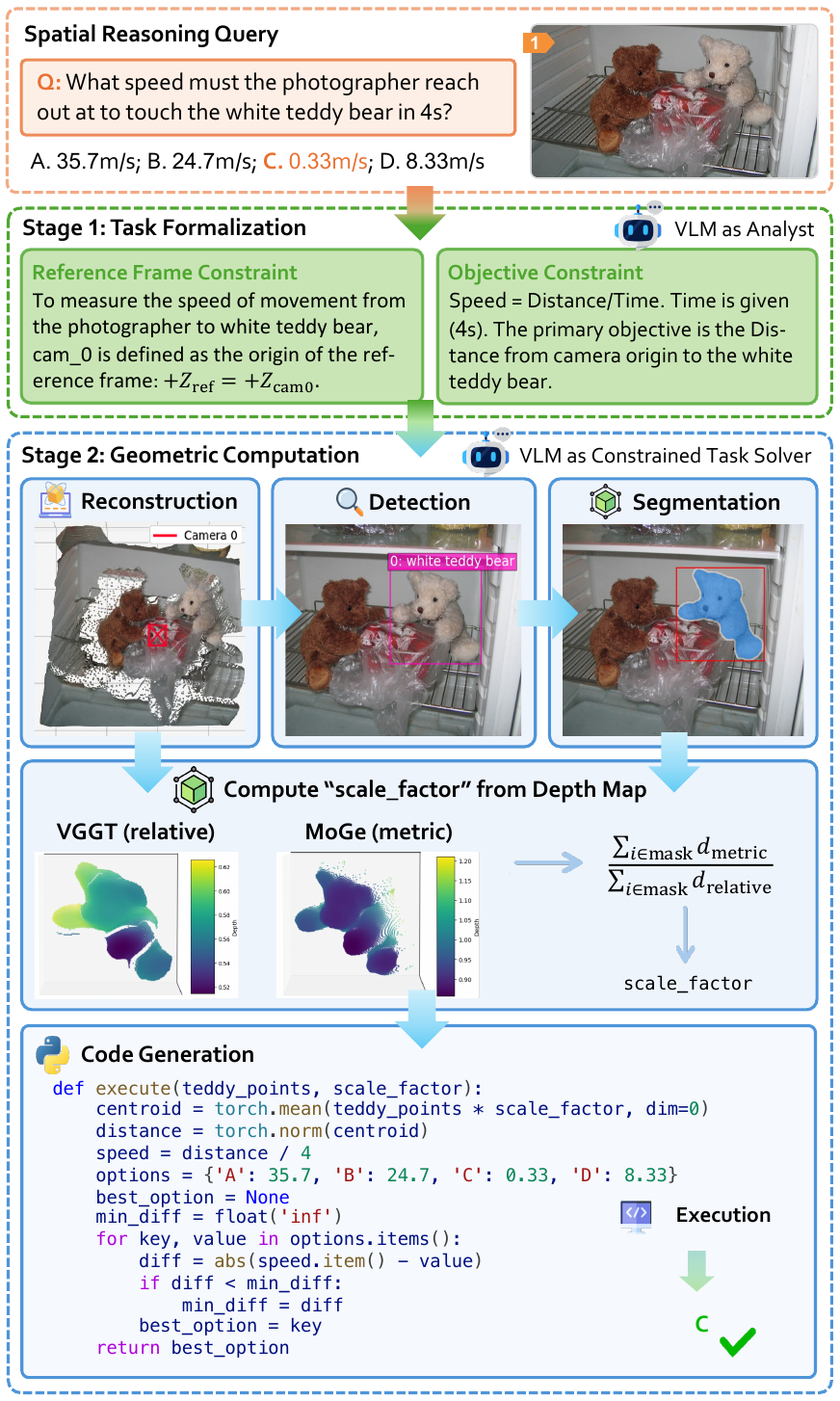}}
\caption{
    \textbf{Case Study \#6.}
    Metric-scale estimation.
}
\label{fig:app_case6}
\vspace{-2em}
\end{center}
\end{figure*}

We provide several qualitative case studies on how \METHOD{} effectively tackles spatial reasoning queries.
These challenging cases includes unique object counting across multiple views (\Cref{fig:app_case1}), direction-based reference frame (\Cref{fig:app_case2}), object-based reference frame (\Cref{fig:app_case3}), camera rotation analysis (\Cref{fig:app_case4}), object movement analysis (\Cref{fig:app_case5}), and metric-scale estimation (\Cref{fig:app_case6}).

\section{Broader Impacts}
\label{sec:app_broader_impacts}

\textbf{Advancing Embodied AI Systems.} 
Applications in robotics and AR/VR depend on an agent's ability to translate ambiguous human instructions (e.g., ``sit on the sofa'') into precise geometric actions.
\METHOD{} provides a robust framework for this translation, potentially facilitating the development of robots and AR/VR interfaces that can interact with the physical world with high fidelity.

\vspace{+0.5mm}
\noindent\textbf{Trust, Interpretability, and Verification.}
Unlike opaque end-to-end spatial VLMs, \METHOD{}'s reasoning process is highly traceable. 
$\gC_\text{task}$ serves as an explicit, human-readable artifact that can be verified before a high-stakes action is executed. 
This verify-then-execute capability is critical for safety in real-world applications. 
Furthermore, the structured outputs from both the $\gF_\text{formalize}$ and $\gF_\text{compute}$ stages can serve as a valuable source of process-level supervision, acting as a reliable validator to guide the training of more efficient end-to-end spatial VLM.

\vspace{+0.5mm}
\noindent\textbf{Inheritance and Amplification of Bias.}
The reliance on VFMs for perception and geometry (\eg{}, 3D reconstruction, object orientation) creates a new dependency chain for bias and failure.
If these perceptual tools perform poorly on objects, scenes, or lighting conditions, \METHOD{} will not only inherit these biases but may also amplify them, leading to incorrect outcomes in real-world interactions.

\clearpage

\begin{table*}[t]
\centering
\caption{
    \textbf{Prompts Used for Formalizing Reference Frame Constraint.}
    Here, \texttt{\{example\}} is the in-context examples (see~\Cref{fig:app_in_context}), and \texttt{\{question\}} is the placeholder that will be replaced.
}
\vspace{-0.5em}
\begin{minipage}{0.99\linewidth}
\begin{tcolorbox} [colback=white]
    \small
    
    \textbf{[CORE MISSION]} \\
    You are an expert spatial reasoning analyst. Your sole mission is to analyze a user's question and define the final \textbf{Reference Frame}. Your goal is to find the single element (an object, a camera, or a vector) that provides the ultimate, non-negotiable definition for absolute directions (e.g., North, South) or relative perspectives (e.g., ``front'', ``left'') in the final answer. Ask yourself: \textit{``What element holds the final authority on what `north', `front', or `left' means in the question?''} \vspace{+2mm} \\
    \textbf{[OUTPUT FORMAT]} \\
    Your response \textbf{MUST} be a single, valid JSON object: \vspace{-2.5mm} 
    \begin{alltt}```json
\{
  "reasoning": "A brief, step-by-step logical deduction, explaining WHY this anchor 
                is the arbiter of direction.",
  "formalization": "The precise mathematical mapping of a semantic direction to one 
                    of the \textbf{Solvable Geometric Primitive} listed below, e.g., 
                    $-Z_\text{cam0}, +Z_\text{toaster}$."
\}
```\end{alltt} 
    \vspace{-1.5mm} \textbf{[FORMALIZATION]} \\
    \textbf{1. Identify the Final Arbiter of Direction} \\
    - \textbf{Priority 1: Absolute Direction.} If an absolute direction (North, South, etc.) is explicitly tied to an element, that element is the arbiter, overriding everything else. \\
    - \textbf{Priority 2: Relative Query.} If no absolute direction is given, the arbiter is the object of the relative question. \vspace{+0.5mm} \\
    \textbf{2. Formalize Reference Frame Using Solvable Geometric Primitives}: o construct a mathematical formalization of reference frame, you MUST use following three types of \textbf{Solvable Geometric Primitives}: \\
    - \textbf{A. Camera Axes}: A vector from a specific camera's coordinate system. Format: $\pm X_\text{cam[i]}, \pm Y_\text{cam[i]}, \pm Z_\text{cam[i]}$. The camera coordinate system follows OpenCV convention: +Z points forward, +Y points down, and +X follow right-hand rules. \\
    - \textbf{B. Object Axes}: A vector from a specific object's semantic coordinate system. Format: $\pm X_\text{[obj]}, \pm Y_\text{[obj]}, \pm Z_\text{[obj]}$. The object's local coordinate system is defined by: +Z points its semantic ``front'', +Y points its semantic ``down'', +X follows right hand rules, and origin at centroid. \\
    - \textbf{C. Inter-Object Vector (Direction)}: A vector connecting the centroids of two concrete, detectable objects. Format: $\text{Centroid(B)}-\text{Centroid(A)}$. \vspace{+0.5mm} \\
    \textbf{3. Semantic Formalization} \\
    - \textbf{A. Object-based Reference Frame}: Usually can be defined by corresponding object's axes. Examples: \\
        \hspace*{+3mm} - ``when using the toaster'' suggests user's ``forward'' is opposite the toaster's semantic ``front'', i.e., $+Z_\text{ref} = -Z_\text{toaster}$. \\
        \hspace*{+3mm} - You must choose a \textbf{Physical and Detectable} object as the object anchor. Don't use abstract concepts like room/region/area.
    - \textbf{B. Camera-based Reference Frame}: Usually can be defined by corresponding camera's axes. Examples: \\
        \hspace*{+3mm} - ``from the perspective of Figure 1'' suggests reference frame is identical to camera 0's, i.e., $+Z_\text{ref} = +Z_\text{cam0}$. \\
    - \textbf{C. Direction-based Reference Frame} \\
        \hspace*{+3mm} - For spatial relationship between two \textbf{Physical and Detectable Objects}, it can be defined by inter-object vector. Examples: ``object A is north of object B'' suggests the direction from object B to A is north, i.e., $+Z_\text{ref} = \vec{{BA}} = \text{Centroid(A)} - \text{Centroid(B)} = \text{North}$ \\
        \hspace*{+3mm} - For spatial relationship between two \textbf{Abstract Concepts}, you must use a physic object's axes or a camera's axes as the proxy to tie this direction. Examples: ``moves from room A to room B, facing north''.Assume this motion aligns with moving from background towards the foreground, formalized as $+Z_\text{ref} = -Z_\text{cam[i]} = \text{North}$. \vspace{+2mm} \\
    \textbf{[EXAMPLES]} \\
    \{examples\} \vspace{+2mm} \\
    \textbf{[QUESTION]} \\
    \{question\} \vspace{+2mm} \\
    Now, please analyze the above question and provide your response in the specified JSON format.
\end{tcolorbox}
\label{tab:app_prompt_formalize_reference}
\end{minipage}
\end{table*}

\begin{table*}
\centering
\caption{
    \textbf{Prompts Used for Formalizing Objective Constraint.}
    Here, \texttt{\{question\}} is the placeholder that will be replaced.
}
\vspace{-0.5em}
\begin{minipage}{0.99\linewidth}
\begin{tcolorbox} [colback=white]
    \small
    
    \textbf{[CORE MISSION]} \\
    You are an expert spatial reasoning analyst. Your sole mission is to analyze a user's question and define the final \textbf{Objective}. Your goal is to rephrase the user's natural-language question into a single and precise sentence. This sentence describes the specific value or piece of information that definitively answer the question. Ask yourself: \textit{``What is the single, final piece of information (e.g., a scalar value, a 3D vector, a sequence of rotations) that the user is finding?''} \vspace{+2mm} \\
    \textbf{[OUTPUT FORMAT]} \\
    Your response \textbf{MUST} be a single, valid JSON object: \vspace{-2.5mm} 
    \begin{alltt}```json
\{
  "reasoning": "A brief, step-by-step logical deduction that breaks down the user's 
                question into its final objective.",
  "formalization": "A single, concise sentence that defines the ultimate goal of the 
                    question, stated in technical terms."
\}
```\end{alltt}
    \vspace{-1.5mm} \textbf{[OBJECTIVE]} \\
    - \textbf{Identify the target variable}: What type of answer is being sought? Is it a distance, a speed, an orientation, a direction, a count, a relationship, or a sequence of actions? \\
    - \textbf{Identify the Key Entities}: What are the specific, concrete objects, cameras, or locations involved in the question? \\ 
    - \textbf{Synthesize the Objective}: Combine the target variable and entities into a single, unambiguous sentence. This sentence must be a statement or a noun phrase, not a question. Example: \\
        \hspace*{+3mm} - Bad: Which way the object are going? \\
        \hspace*{+3mm} - Good: The 3D direction vector of the object' movement. \vspace{+2mm} \\
    \textbf{[QUESTION]} \\
    \{question\} \vspace{+2mm} \\
    Now, please analyze the above question and provide your response in the specified JSON format.
\end{tcolorbox}
\label{tab:app_prompt_formalize_objective}
\end{minipage}
\end{table*}

\begin{table*}[t]
\centering
\caption{
    \textbf{Prompts Used for Tool Orchestration.}
    Here, \texttt{\{api\_documents\}} is the detailed documentation of provided APIs. 
    \texttt{\{history\}} includes the initial user question, task formalization, previous planning and corresponding execution results.
}
\vspace{-0.5em}
\begin{minipage}{0.99\linewidth}
\begin{tcolorbox} [colback=white]
    \small
    
    \textbf{[CORE MISSION]} \\
    You are an expert spatial intelligence agent. Your mission is to generate a sequence of tool calls that rigorously computes the answer. It follows an iterative Plan $\rightarrow$ Update cycle, using a workspace as your computational memory. \\
    \hspace*{+3mm} - \textbf{Plan}: Decide the next tool calls based on the goal and current workspace. \\
    \hspace*{+3mm} - \textbf{Update}: Tool results are saved as new variables in the workspace. \\
    \hspace*{+3mm} - \textbf{Repeat}: Continue until the workspace contains enough information to conclude the final answer. \vspace{+2mm} \\
    \textbf{[AVAILABLE APIS]} \\
    \{api\_documents\} \vspace{+2mm} \\
    \textbf{[A TYPICAL WORKFLOW]} \\
    \textbf{1. Strictly Follow the Task Formalization} \\
    \hspace*{+3mm} - \textbf{Task Formalization}: The input question is pre-formalized and consists of two parts: \textbf{Reference Frame Constraint} and \textbf{Objective Constraint}. You MUST strictly follow this formalization to solve the question. \\
    \hspace*{+3mm} - \textbf{Reference Frame}: Reference frame is the only one coordinate system that matters for interpreting the final answer (left/right, north/south, etc.). The ``formalization'' is the equation you must solve with the specified geometric tools. E.g., $+Z_\text{ref} = -Z_\text{toaster}$ indicates reference frame is defined by object toaster's local frame, so we MUST perform all calculations in toaster's frame. \\
    \hspace*{+3mm} - \textbf{Objective}: Objective is the ultimate goal of the question. You must calculate this objective within the reference frame. \vspace{+0.5mm} \\
    \textbf{2. Acquire Geometric Data}: Based on user's question and pre-defined task formalization, plan the necessary tool calls to gather all data required for the final calculation. This involves two parallel goals: \\
    - \textbf{A. Solve for the Reference Frame}: The formalization mathematically defines the \textbf{World-to-Reference Transformation}. To solve this formalization, your plan MUST gather all the geometric ingredients in the world frame. \\
        \hspace*{+3mm} - ``reconstruct'' tool provides the 3D reconstruction context in a unified world frame, bridging the gap between input 2D images and geometric perception. \\
        \hspace*{+3mm} - If formalization involves an object's axes (e.g., $+Z_\text{ref} = -Z_\text{toaster}$), call ``predict\_obj\_pose'' to solve that object's local frame. The resulting ``T\_obj2world'' is required to establish the reference frame. \\
        \hspace*{+3mm} - If formalization involves a camera's axes (e.g., $+Z_\text{ref} = -Z_\text{cam0}$), you must acquire reconstruction context (include extrinsic matrix) to implement the formalization. \\
        \hspace*{+3mm} - If formalization involves a vector between two objects (e.g., $\text{Centroid(B)} - \text{Centroid(A)} = \text{North}$), your plan MUST include calls to ``project\_box\_to\_3d\_points'' for both object A and B. \\
    - \textbf{B. Solve for the Objective}: Follow the objective to identify the target data that need to be analyzed within the reference frame.\vspace{+0.5mm} \\
    \textbf{3. Perform Final Calculation in Reference Frame}: Once all required variables are available in the workspace, call ``code''. \vspace{+0.5mm} \\
    \textbf{4. Conclusion}: Conclude the final answer using ``generate\_final\_answer''. \vspace{+2mm} \\
    \textbf{[OUTPUT FORMAT]} \\
    Your response \textbf{MUST} be a single, valid JSON object: \vspace{-2.5mm}
    \begin{alltt}```json
\{
  "analysis": "Briefly analyze how you will implement the formalization and what 
               target data is need. State the immediate next tool(s) you will call",
  "tool_calls": [
    \{
      "api": "API name", "args": \{...\},
      "output_variable": "A unique name for output, stored in the workspace"
    \}, ...
  ]
\}
```\end{alltt}
    \vspace{-1.5mm} \textbf{[HISTORY]} \\
    Here is the history so far: \\
    \{history\} \vspace{+2mm} \\
    Please analyze current situation and history messages, and then generate your response. Your plan MUST only includes the immediate next one step.
\end{tcolorbox}
\label{tab:app_prompt_tool}
\end{minipage}
\end{table*}

\begin{table*}[t]
\centering
\caption{
    \textbf{Prompts Used for Coder.}
    Here, \texttt{\{question\}}, \texttt{\{formalization\}}, \texttt{\{objective\}} and \texttt{\{var\_docs\}} are the placeholder that will be replaced.
    \texttt{\{knowledge\}} includes a set of releveant, fixed formulas based on the type of input variables.
}
\vspace{-0.5em}
\begin{minipage}{0.99\linewidth}
\begin{tcolorbox} [colback=white]
    \small
    
    \textbf{[CORE MISSION]} \\
    You are an expert Python programmer. Your goal is to write a single Python function that correctly implements the computational objective based on the provided context and documentation. \vspace{+2mm} \\
    \textbf{[User's Question]} \\
    This provides the high-level context for your task. \\
    \{question\} \vspace{+2mm} \\
    \textbf{[Reference Frame]} \\
    All geometric data are defined in the world frame (defined by camera 0) unless specified. All final interpretations MUST be expressed in the reference frame. The reference frame is defined: \\
    \{formalization\} \vspace{+2mm} \\
    \textbf{[Objective]} \\
    The ultimate goal from the high-level question. You MUST write code to calculate this objective to answer the question. \\
    \{objective\} \vspace{+2mm} \\
    \textbf{[Documentation of Available Variables]} \\
    \{var\_docs\} \vspace{+2mm} \\
    \textbf{[Additional Knowledge]} \\
    This information is always true for the environment your code runs in. \\
    \{knowledge\} \vspace{+2mm} \\
    \textbf{[Available Libraries]} \\
    You can use ``numpy'', ``torch'', ``scipy'', ``math'', and other standard Python libraries. \\
    \textbf{[Critical Rules and Output Format]} \\
    \textbf{1. Synthesize and Self-Correct}: Your primary duty is to write correct code. Use the objective as your goal, but critically verify and implement the logic using the provided documentation. \\
    \textbf{2. Handle Multiple-Choice Questions}: If the user's question is multiple-choice, your code MUST systematically evaluate the conditions for every option (e.g., A, B, C, D). Besides, the logic of your code should focus ONLY on the given options. \\
    \textbf{3. DO NOT add any explanation in your final output.} Your output MUST follow this format: \vspace{-2.5mm}
    \begin{alltt}```python
def execute({func_signature}):
    # Your code here
    ...

    return serializable_value  # return value MUST be a serializable type
```\end{alltt}
\end{tcolorbox}
\label{tab:app_prompt_code}
\end{minipage}
\end{table*}

\begin{table*}[t]
\centering
\caption{
    \textbf{Knowledge and Formulas Used in Knowledge-Augmented Code Generation.}
    We will inject the relevant knowledge based on the type of input variable.
}
\vspace{-0.5em}
\begin{minipage}{0.99\linewidth}
\begin{tcolorbox} [colback=white]
    \small

    \textbf{[Output of ``reconstruct'']} \\
    \textbf{Extrinsic Transformation (World $\leftrightarrow$ Camera)}: \\
    - \textbf{World $\rightarrow$ Camera}: 
    To transform a world point ``P\_world'' into camera s's frame, use its extrinsic matrix \texttt{E\_s = extrinsic[s]}. 
    The formula is: \texttt{P\_cam\_homo = P\_world\_homo @ E\_s.T}. \\
    - \textbf{Camera $\rightarrow$ World}: 
    The pose of camera ``s'' in the world, ``Pose\_s'', is the inverse of its extrinsic matrix: 
    \texttt{Pose\_s = np.linalg.inv(extrinsic[s])}. 
    To transform a point from camera s's local frame to the world frame, use the formula: 
    \texttt{P\_world\_homo = P\_cam\_homo @ Pose\_s.T}. \\
    - \textbf{Relative Rotation Analysis (Camera $\rightarrow$ Camera)}: 
    To describe the rotation of camera j's pose relative to camera i's pose, use the camera poses in the world frame (``Pose\_i'', ``Pose\_j''). 
    The relative rotation from i to j is: 
    \texttt{R\_rel = R\_j\_pose @ R\_i\_pose.T}, 
    which simplifies to \texttt{R\_rel = (R\_j.T) @ (R\_i.T).T = R\_j.T @ R\_i}.
    
    \noindent\makebox[\linewidth]{\tikz[baseline]{\draw[dashed] (0,0) -- (\linewidth,0);}}

    \vspace{+1mm} \textbf{[Output of ``predict\_obj\_pose'']} \\
    \textbf{Object Pose Transformation (World $\leftrightarrow$ Object)}: \\
    - \textbf{Object $\rightarrow$ World}: 
    The ``T\_obj2world'' matrix (aliased as ``Pose\_obj'') transforms points from the object's local frame to the world frame using the formula:
    \texttt{P\_world\_homo = P\_local\_homo @ Pose\_obj.T}. \\
    - \textbf{World $\rightarrow$ Object}: 
    To transform ``P\_world'' into the object's local frame, use the inverse matrix: 
    \texttt{T\_world\_to\_obj = np.linalg.inv(Pose\_obj)}. 
    The formula is: \texttt{P\_local\_homo = P\_world\_homo @ T\_world\_to\_obj.T}.

    \noindent\makebox[\linewidth]{\tikz[baseline]{\draw[dashed] (0,0) -- (\linewidth,0);}}

    \vspace{+1mm} \textbf{[Output of ``reconstruct'', ``predict\_obj\_pose'']} \\
    \textbf{Interpreting Rotation}: \\
    - \textbf{General Rotation Direction}: 
    For most questions about rotation direction, convert the relative rotation matrix to a rotation vector \texttt{[rx, ry, rz]} via \texttt{scipy.spatial.transform.Rotation.from\_matrix(R).as\_rotvec()}. \\
        \hspace*{+3mm} - The component with the largest absolute value indicates the primary axis of rotation. \\
        \hspace*{+3mm} - Based on the OpenCV coordinate system (+X right, +Y down, +Z forward) and the right-hand rule:
        \texttt{ry > 0} corresponds to a pan to the right, \texttt{rx > 0} corresponds to a tilt upward, \texttt{rz > 0} corresponds to a clockwise roll. \\
    - \textbf{Sequential Rotations}: 
    Use \texttt{scipy.spatial.transform.Rotation.from\_matrix(R).as\_euler(order)} to compute sequential rotation. Verify the signs of the resulting angles. 
    It must match the option's description.

    \noindent\makebox[\linewidth]{\tikz[baseline]{\draw[dashed] (0,0) -- (\linewidth,0);}}

    \vspace{+1mm} \textbf{[If formalization includes cardinal direction]} \\
    \textbf{1. Identify the Cardinal Anchor Axis from the formalization}. \\
        \hspace*{+3mm} - The formalization string (e.g., $+Z_\text{ref} = -Z_\text{obj} = \text{South}$) links one of your reference axes to a cardinal direction. \\
        \hspace*{+3mm} - Parse this string to find the anchor. 
        In the example, the anchor is South, and it corresponds to the $Z_\text{ref\_axis}$. 
        This defines your first cardinal vector: $\text{South\_axis} = Z_\text{ref\_axis}$. \\
    \textbf{2. Derive the Complete Set of Cardinal Axes}: You must find the remaining cardinal axes via applying cross product: \\
        \hspace*{+3mm} - If South\_axis is known, starting with \texttt{West\_axis = np.cross(Y\_ref\_axis, South\_axis)}. \\
        \hspace*{+3mm} - If West\_axis is known, starting with \texttt{North\_axis = np.cross(Y\_ref\_axis, West\_axis)}. \\
        \hspace*{+3mm} - If North\_axis is known, starting with \texttt{East\_axis = np.cross(Y\_ref\_axis, North\_axis)}. \\
        \hspace*{+3mm} - If East\_axis is known, starting with \texttt{South\_axis = np.cross(Y\_ref\_axis, East\_axis)}. \\
    \textbf{3. Project and Determine the Final Quadrant}: 
    Project the target vector onto the primary horizontal cardinal axes (North and East). \\
        \hspace*{+3mm} - \texttt{projection\_north = np.dot(disp\_vec\_world, cardinal\_map["N"])} \\
        \hspace*{+3mm} - \texttt{projection\_east = np.dot(disp\_vec\_world, cardinal\_map["E"])} \\
        \hspace*{+3mm} - Use the signs of these projections to determine the final answer.

    \noindent\makebox[\linewidth]{\tikz[baseline]{\draw[dashed] (0,0) -- (\linewidth,0);}}

    \vspace{+1mm} \textbf{[Output of ``detect'']} \\
    \textbf{Bounding Box Format}: All bounding box in input variable is provided in the \texttt{[x1, y1, x2, y2]} format.
    
\end{tcolorbox}
\label{tab:app_prompt_kacg}
\end{minipage}
\end{table*}

\end{document}